\documentclass[journal,draftcls,onecolumn,12pt,twoside]{IEEEtran}

\usepackage[T1]{fontenc}
\ifCLASSINFOpdf

\else
\fi

\usepackage[pdftex]{graphicx}
\usepackage{graphicx}
\usepackage{caption}
\usepackage{subcaption}
\usepackage{amsmath,amsthm}
\usepackage{optidef}
\usepackage[cmintegrals]{newtxmath}
\usepackage{algorithmic}
\usepackage[vlined,ruled,commentsnumbered,linesnumbered]{algorithm2e}
\usepackage{setspace}

\newcounter{mytempeqncnt}

\newtheorem{thm}{Theorem}
\theoremstyle{definition}
\newtheorem{definition}{Definition}
\theoremstyle{proposition}

\newtheorem{lem}{Lemma}
\newtheorem{rem}{Remark}

\begin{document}

\title{Decentralized Learning of Tree-Structured Gaussian Graphical Models from Noisy Data}

\author{Akram Hussain
\thanks{The authors are with the Department of Computer Science and Engineering, Shanghai Jiao Tong University, Shanghai, China (e-mail: akram.hussain@iba-suk.edu.pk)

}
}

\maketitle

\begin{abstract}
This paper studies the decentralized learning of tree-structured Gaussian graphical models (GGMs) from noisy data. In decentralized learning, data set is distributed across different machines (sensors), and GGMs are widely used to model complex networks such as gene regulatory networks and social networks. The proposed decentralized learning uses the Chow-Liu algorithm for estimating the tree-structured GGM.

In previous works, upper bounds on the probability of incorrect tree structure recovery were given mostly without any practical noise for simplification. While this paper investigates the effects of three common types of noisy channels: Gaussian, Erasure, and binary symmetric channel. For Gaussian channel case, to satisfy the failure probability upper bound $\delta > 0$ in recovering a $d$-node tree structure, our proposed theorem requires only $\mathcal{O}(\log(\frac{d}{\delta}))$ samples for the smallest sample size ($n$) comparing to the previous literature \cite{Nikolakakis} with $\mathcal{O}(\log^4(\frac{d}{\delta}))$ samples by using the positive correlation coefficient assumption that is used in some important works in the literature. Moreover, the approximately bounded Gaussian random variable assumption does not appear in \cite{Nikolakakis}. Given some knowledge about the tree structure, the proposed Algorithmic Bound will achieve obviously better performance with small sample size (e.g., $< 2000$) comparing with formulaic bounds. Finally, we validate our theoretical results by performing simulations on synthetic data sets.
\end{abstract}

\begin{IEEEkeywords}
Structure learning, Chow-Liu algorithm, tree-structured Gaussian graphical model, noisy data set.
\end{IEEEkeywords}

\IEEEpeerreviewmaketitle

\section{Introduction}\label{sectionI}

\IEEEPARstart{M}{odern} learning systems are distributed in nature due to the lack of capacity of current storage systems. Two types of systems are very common in machine learning: decentralized (having a fusion center (FC)) and fully distributed systems. In decentralized systems, all the local machines are connected to the FC via communication channels. \textbf{In a decentralized system, machines generally take measurements and forward their decision to the FC that performs estimation or learning tasks, depending on the goal} \cite{Xiao}. On the other hand, machines process data cooperatively without a central server (FC) in a fully distributed system.

Moreover, data is distributed across different geographical locations in two ways: the horizontally distributed data and vertically distributed data. In the horizontally distributed setting, some samples of the data set with all the dimensions (components) are given to each machine, however, in the vertically distributed setting, one dimension (component) of all the samples of the data set is usually allocated to a machine.

\subsection{Motivation and Literature Review}
To study complex phenomena (e.g., gene selection in cancer classification \cite{YWang}) by using big data requires us to characterize the relationships among a large number of variables. In this paper, our focus is on decentralized learning of tree-structured Gaussian graphical models (GGMs) from noisy vertically distributed data set. The noise in the data set is due to the noisy communication between sensors and the FC, the limited power of the sensors, environmental harsh conditions of operation, and straggler sensors: sensors that are significantly slower than the average \cite{Lee}, \cite{Buyukates}.

Tree-structured GGM is the special model of unidrected graphical models (UGMs) that are important tools for tractable modeling of multivariate distributions. Furthermore, the factorization properties, associated with the graphical models, are less complex computationally. UGMs are Markov random fields (MRFs), and these models are more natural for some problems such as image analysis and spatial statistics \cite{Murphy}, \cite{Bishop}. If the multivariate distributions are also assumed to be Gaussian, then the models will be called GGMs. GGMs are widely used for complex networks such as gene selection for cancer classification \cite{YWang}, brain connectivity networks \cite{SHuang}, and social networks \cite{RXiang}. 

Structure learning: finding the allowed dependencies among variables, is a model selection problem where a graph is chosen that corresponds to the dependence structure in a given data set. If the graph is tree structure, the efficient maximum likelihood (ML) method is given by Chow-Liu \cite{Chow} to estimate the probability distribution from the data set. The ML estimator is equivalent to maximum weight spanning tree (MWST) problem which can efficiently be solved by Kruskal \cite{Kruskal} or Prim algorithm \cite{Prim}. \textbf{The tree-structured models are practically utilizable because a small sample size and low time complexity are required as compared to graphs with loops for computing the ML \cite{Bresler}}. The error exponent of Chow-Liu algorithm is analyzed in \cite{Tan, F.Tan}, and \cite{Tandon} for tree-structured discrete distribution, continuous distribution, and tree-structured Ising model with side information, respectively.
The tree-structured graphical models are learned in the decentralized system \cite{Tavass} with Gaussian random variables and the fully distributed system \cite{Jang} with binary random variables under communication constraints. Tavassolipour \textit{et al.} study the structure learning problem where the data set is distributed vertically across machines \cite{Tavass}.

The inverse covariance matrix estimation in GGMs is studied with the known graphical model \cite{Wiesel, Meng} in the distributed environment where distributed machines collaboratively estimate the inverse covariance matrix using message passing protocols. Tavassolipour \textit{et al.} \cite{A.Tavass} study the structure learning of GGM (not tree structure): inferring the non-zero pattern of the precision matrix from the vertically distributed data set. 

Structure learning has also been performed from noisy data set such as a general graph is learned in \cite{Kang} from the noisy data set. In addition, the tree-structured GGM recovery is studied in \cite{Katiyar} from Gaussian corrupted noisy data set. Similarly, Katiyar \textit{et al.} studies the problem of recovering tree-structured Ising model from non-identically distributed noises \cite{Katiyar2}. The sample complexity requirement for learning tree-structured Ising model and GGM is studied under reliable data set \cite{Bresler} and noisy data set \cite{Nikolakakis}, \cite{Nikolakakis1} based on Chow-Liu algorithm. In addition, it has been proved in \cite{Nikolakakis2} that Chow-Liu algorithm gives the ML estimate of the tree for noisy data set where the noise is not necessarily identically distributed. For hidden non-parametric tree structure learning of graphical model, they also provide the lower bounds on the sample complexity for the general alphabet (i.e., countable and finite). Partial recovery of tree-structured graphical model is studied in \cite{Tandon2} from noisy data set where the noise is non-identically distributed.

\subsection{Our Contributions}
This paper studies the decentralized learning of tree-structured GGMs from noisy data sets that are corrupted by Gaussian, Erasure, and binary symmetric channel (BSC) noises. Moreover, the proposed results are simulated using synthetic data sets. The main contributions of this paper are as follows.
\begin{itemize}
  \item \textbf{For Nonquantized Data Set:} Theorem \ref{theorem1} and Theorem \ref{theorem2} provide the upper bounds on the probability of incorrect recovery of tree-structured GGMs in the presence of (noisy) Gaussian and Erasure channels between sensors and the FC with finite size data set, respectively. These results are obtained using the positive correlation coefficient assumption that has been used previously for learning Ising model \cite{Murphy}, \cite{Tandon} and associative Markov networks \cite{Taskar}. For Gaussian channels, the proposed result needs only $\mathcal{O}(\log(\frac{d}{\delta}))$ samples for the smallest sample size ($n$) as compared with \cite{Nikolakakis} that requires $\mathcal{O}(\log^4(\frac{d}{\delta}))$ samples.
  \item \textbf{For Quantized Data Set:} We extend the work of \cite{Tavass} by studying the impact of (noisy) BSC on the tree structure learning performance, where BSC can be regarded as the quantized Gaussian. An upper bound on the probability of a crossover event is derived in Lemma \ref{lemma4} for the noisy quantized data set. Furthermore, it is shown empirically that the channel noise heavily degrades the learning performance even for a small tree.
  \item \textbf{Algorithmic Bound:} When some knowledge about the tree structure is known, Algorithmic Bound (Algorithm \ref{algorithm2}) is proposed especially for a small data set. The proposed Algorithmic bound is theoretically in Theorem \ref{theorem3} and empirically analyzed, and  it achieves obviously better performance with the small data set (e.g., $< 2000$) because it utilizes the external knowledge as side information as compared to Theorem \ref{theorem2} for Erasure channel.
\end{itemize}


\subsection{Paper Organization and Notations}
The paper is organized as follows. Section \ref{sectionII} states GGMs and problem statement. Some main results are presented in Section
\ref{sectionIII}-\ref{sectionIV} for Gaussian, Erasure, and BSC noises. Algorithmic Bound is proposed in Section \ref{sectionV} to tight the performances of Theorem \ref{theorem1} and Theorem \ref{theorem2}. In addition, the experiments are performed to validate the proposed bounds in Section \ref{sectionVI}, and the paper is concluded in Section \ref{sectionVII}.

Our proposed bounds are evaluated on synthetic data sets, and similar data sets have been used in literature such as \cite{F.Tan}-\cite{Jang}, \cite{Nikolakakis}.

We use capital letters (e.g., $X$) for random variables and small letters for their realizations. The vectors and their realizations are represented by bold-faced capital and small letters, respectively. For example, random variables are denoted by $\hat{I}_e$, $L, O,$ and $T$.
With the exception, estimator of the correlation coefficient $\hat{\rho}_e$ and the parameter $\hat{\theta}_{ij}$ are also random variables. In addition, the symbols $\lambda$, $\alpha$, and $t$ are small positive constants, and $\mathcal{I}(\cdot)$ and $\mathbb{E}[\cdot]$ represent an indicator function and the expectation operator, respectively. Moreover, words such as machines and sensors are interchangeably used in this paper.

\section{Preliminaries and Problem Statement}\label{sectionII}
\subsection{Gaussian Graphical Models (GGMs)}
UGMs represent probability distributions that factorize according to the structures of given undirected graphs \cite{Drton}, \cite{Tan}, \cite{F.Tan}, \cite{Wiesel}, \cite{Meng}.
Let $G = (V, E)$ be an \textbf{undirected simple graph} with the vertex set $V = [d] = \{1, 2, ..., d\}$ and edge set $E \subset \binom{V}{2}$, where the vertex set indexes the random variables and the dependence constraints are encoded by the edge set.

Given a $d-$dimenstional random vector $\textbf{X} = (X_1, ..., X_d)^T$, the distribution $p(\textbf{x})$ satisfies the local Markov property on a graph $G = (V, E)$ if
$$p(x_i | x_{V\diagdown\{i\}}) = p(x_i | x_{N(i)}),\quad \forall i \in V,$$
where $N(i)$ is the set of neighbors of node $i$ and is defined as $N(i) := \{j \in V: (i, j) \in E\}$.

In this paper, it is assumed that the multivariate distribution $p$ is a GGM with mean zero and unknown positive definite covariance matrix $\Sigma \succ 0$, and moreover, it is also assumed that the distribution $p$ is Markov on tree graph $T_p := G$ (i.e., $T_p$ represents the graphical model with distribution $p$ on tree $T$). Since probability distribution $p$ is Markov on tree $T_p$, it has the following factorization property
\begin{equation}\label{eq:1}
p(\textbf{x}) = \prod_{i\in V} p_i(x_i) \prod_{(i, j) \in E}\frac{p_{i, j}(x_i, x_j)}{p_i(x_i)p_j(x_j)},
\end{equation}
where $p_i(x_i)$ is the node marginal and $p_{i, j}(x_i, x_j)$ is the pairwise marginal for each node $i \in V$ and for each pair $(i, j) \in E$, respectively. Due to GGM assumption, the $p(\textbf{x})$ can be written as
\begin{equation}\label{eq:2}
p(\textbf{x}) = \frac{1}{(2\pi)^{d/2}|\Sigma|^{1/2}} exp(-\frac{1}{2}\textbf{x}^T \Sigma ^{-1} \textbf{x}).
\end{equation}

\subsection{Problem Statement}
The goal of this paper is to estimate the structure of $T_p$ given $n$ independent $d$-dimensional samples $\{\textbf{x}^{(1)}, ..., \textbf{x}^{(n)}\}$ where the superscript is the time index, for instance, $\textbf{x}^{(k)} = (x_1^{(k)}, ..., x_d^{(k)})$ is the snapshot for time $k$. In addition, $\{x_i ^{(1)}, ..., x_i ^{(n)}\}$ is the data set on machine (sensor) $i$ with time index from $1$ to $n$ which is called vertically distributed data set. The data set $\{\textbf{x}^{(1)}, ..., \textbf{x}^{(n)}\}$ is generated independently from a $d-$dimensional zero mean normal distribution $\mathcal{N(\bf 0, \Sigma)}$ which can be factorized into node and pairwise marginals according to tree $T_p = (V, E)$ using formula (\ref{eq:1}). 

\begin{figure}[t]
\centering
\includegraphics[width=3in]{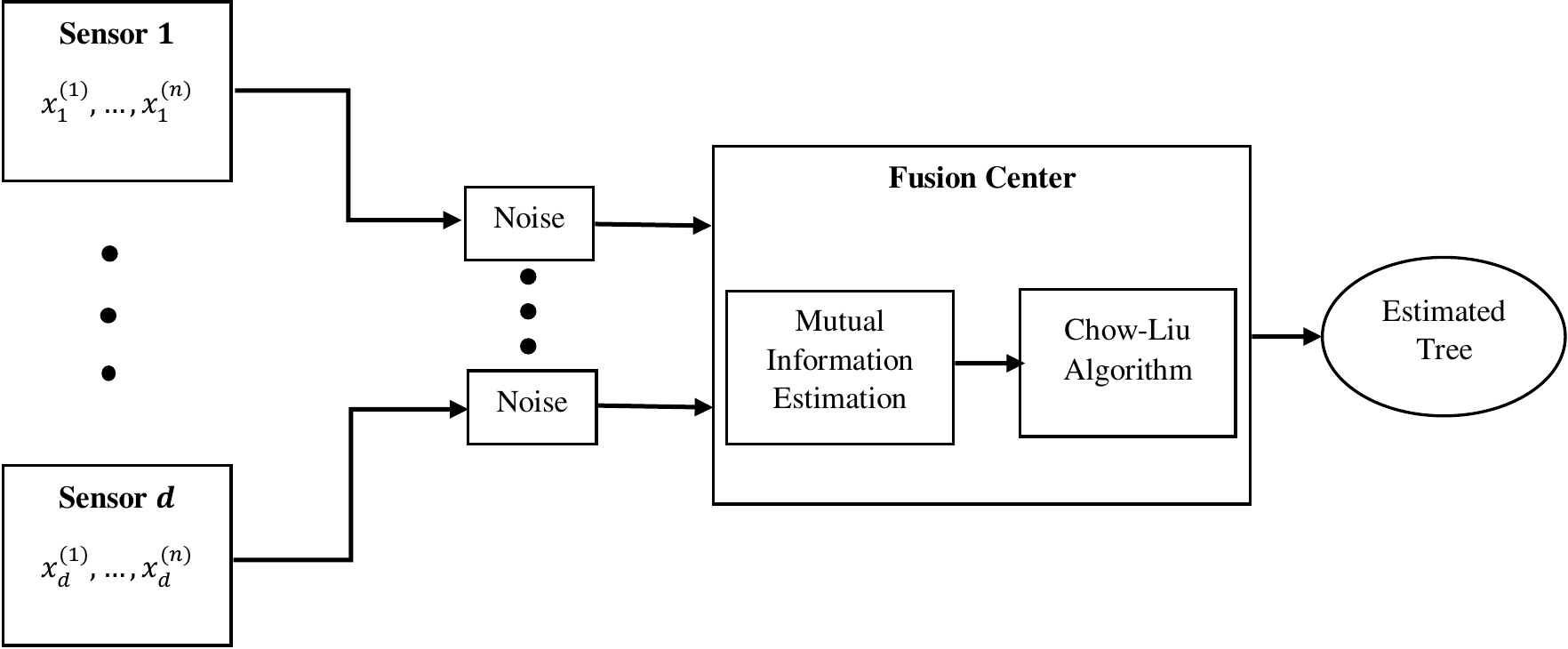}
\caption{Block diagram for decentralized learning of tree-structured GGMs.}
\label{model}
\end{figure}

In \cite{Tavass}, they bound the probability of incorrect tree-structured GGM recovery with finite size data set by assuming the noise free communication between sensors and the FC. However, the communication between sensors and the FC is noisy in general as mentioned in Section \ref{sectionI}. The learning performance is highly dependent on the channel noises. Therefore, we model noisy communications by Gaussian channel, Erasure channel, and BSC in this paper. BSC is used as a special case when the machines quantize the data before transmission to the FC. The FC employs Chow-Liu algorithm (Algorithm 1) to estimate the underlying tree structure when all the data is available to it from all machines as shown in Fig. \ref{model}. 

\subsection{Chow-Liu Algorithm}
The first step in Chow-Liu algorithm is to estimate mutual information between any pair of random variables as shown in Algorithm \ref{algorithm1} and then use Kruskal \cite{Kruskal} or Prim \cite{Prim} algorithm to obtain MWST.
\begin{algorithm}[hbt!]
\LinesNumbered
\SetAlgoLined
\KwIn{Data set: $\{\textbf{x}^{(1)}, ..., \textbf{x}^{(n)}\}$}
\KwOut{Estimated Tree: $\hat{T}_p$}
{

Estimate mutual information for each possible edge : $\hat{I}(X_i, X_j)$, for all $i, j \in V$.

Estimated tree $\hat{T}_p = MaximumWeightSpanningTree\left(\cup_{i \not = j}\{\hat{I}(X_i, X_j)\}\right)$.

\Return $\hat{T}_p$
}
\caption{Chow-Liu Algorithm}
\label{algorithm1}
\end{algorithm}

The mutual information between any pair of random variable, say $X_i$ and $X_j$, is obtained by
\begin{equation}\label{eq:3}
I(X_i; X_j) = -\frac{1}{2}\log(1 - \rho^2_{ij}),
\end{equation}
where $\rho_{ij}$ is the correlation coefficient between $X_i$ and $X_j$. Equation (\ref{eq:3}) is only valid for Gaussian random variables.

Since the data set is given, the mutual information in formula (\ref{eq:3}) has to be estimated. In fact, to obtain an unbiased estimator for mutual information is hard, therefore, one can use the unbiased estimator for $\rho_{ij} ^2$ \cite{Tavass} which is given by
\begin{equation}\label{eq:4}
\tilde{\rho}_{ij}^2 = \frac{n}{n + 1}\left(\hat{\rho}_{ij}^2 - \frac{1}{n}\right),
\end{equation}
where $\hat{\rho}_{ij}$ is the estimator of the correlation coefficient, i.e.,
\begin{equation}\label{eq:5}
\hat{\rho}_{ij} = \frac{1}{n} \sum_{k = 1}^{n}X_i^{(k)}X_j^{(k)}.
\end{equation}


\subsection{Positive Correlation Coefficient Assumption}
A positive correlation coefficient $(\rho > 0)$ assumption has been used previously for learning Ising model \cite{Murphy}, \cite{Tandon} and associative Markov networks \cite{Taskar}. Furthermore, the study \cite{Krumsiek} addresses the reconstruction of metabolic reactions from cross-sectional metabolomics data where they show that the correlation coefficient tends strongly towards positive values in metabolomics data.

GGMs can be used for image analysis \cite{Bishop} with $d$ graph nodes that are correlated. Generally, the correlation between any two nodes is positive, means similar labels are assigned to the nodes with high probability.
Take any two machines $X_i$ and $X_j$ arbitrarily which are correlated positively (e.g., $\rho_{ij} > 0$), and the observation models of these machines are given as $Y_i = X_i + noise_i$ and $Y_j = X_j + noise_j$ where $noise_i$ and $noise_j$ are the i.i.d. random variables representing measurement noises. It is known that $Y_i$ and $Y_j$ are more likely to be positively correlated because $X_i$ and $X_j$ are correlated positively.

After observing the data set, machine $i$ transmits it to the FC through the noisy channel. The received data set at the FC is represented by $Z_i = Y_i + channelNoise_i$. Similarly, for machine $j$, therefore, $Z_i$ and $Z_j$, which represent the data set at the FC, are also positively correlated.

Note: If the correlation coefficients of any two machines $X_i$ and $X_j$ are assumed to be positive (e.g., Associative Markov Networks and Ising Model for ferromagnet), then the estimator of the correlation coefficients of the two machines will also be approximately positive because of arbitrarily small values of random variables $noise$ and $channelNoise$.

\subsection{Related Distribution and Hoeffding's Inequality}
This subsection reviews a normal distribution approximation of normal product density function and Hoeffding's inequality.

\subsubsection{Normal Approximation of Normal Product} Let $X\sim \mathcal{N}(\mu_x,\sigma_x^2)$ and $Y\sim \mathcal{N}(\mu_y,\sigma_y^2)$ be independent random variables, then the distribution of their product $Z = XY$ tends to Normal distribution with parameters (if the skewness goes to zero \cite{Ware}-\cite{Macias}):
\begin{equation}\label{eq:6}
     \begin{split}
     \mu_z & = \mu_x\mu_y, \\
     \sigma_z^2 & = \mu_y^2 \sigma_x^2 + (\mu_x^2 + \sigma_x^2)\sigma_y^2 ,\\
     Skewness & = \frac{6\delta_x \delta_y \sigma_x^3 \sigma_y^3}{((1 + \delta_x^2 + \delta_y^2)\sigma_x^2\sigma_y^2)^{3/2}},
     \end{split}
\end{equation}
where $\delta_x = \frac{\mu_x}{\sigma_x}$ and $\delta_y = \frac{\mu_y}{\sigma_y}$.

\subsubsection{Hoeffding's Inequality} Let $X_1, ..., X_n$ be independent bounded random variables with $X_i \in [a, b]$ for all $i$, where $-\infty < a \leq b < +\infty$. Then
\begin{equation}\label{eq:7}
  Pr(\frac{1}{n}\sum_{i = 1}^{n} (X_i - \mathbb{E}[X_i]) \geq t) \leq exp(-\frac{2nt^2}{(b - a)^2}),
\end{equation}
and
\begin{equation}\label{eq:8}
  Pr(\frac{1}{n}\sum_{i = 1}^{n} (X_i - \mathbb{E}[X_i]) \leq - t) \leq exp(-\frac{2nt^2}{(b - a)^2}),
\end{equation}
for all $t\geq 0$.

\section{Structure Learning in Noisy Communication without Quantization}\label{sectionIII}
In this section, the probability of incorrect recovery of tree-structured GGMs is characterized from the noisy data set. The data set is corrupted by Gaussian and Erasure noises. Moreover, the channel gain matrix of the receiver (e.g., the FC) is assumed to be a unit matrix which implies that there is no fading, however, communication is noisy.

\subsection{Structure Learning in the Presence of Gaussian Channels between Sensors and the Fusion Center (FC)}
In this subsection, the impact of Gaussian channels on the tree structure learning performance is studied. The probability of a crossover event (Definition \ref{def1}) is characterized for the general case in Lemma \ref{lemma2}. In the general case, Gaussian noises of different channels are distributed with different means and the same variances where in general the signal to noise ratio (SNR) is determined by the variance. Moreover, the upper bound on the tree structured incorrect recovery probability is obtained in Theorem \ref{theorem1} for the general case.

Each component (i.e., $X_i$) of a vector representing data is normally distributed having zero mean and unit variance (marginal), i.e., $\mathbb{E}[X_i^2] = 1$, and Gaussian noise random variable has the distribution $N_i \sim \mathcal{N}(0,\sigma^2)$. It is also assumed that the noise variable $N_i$ is independent in time and channels.

Random variable $X_i$ is the generic random variable of a sequence of i.i.d. random variables $\{X_i^{(1)}, ..., X_i^{(n)}\}$ which represent the original data at machine $i$. Similarly, $Y_i$ is the generic random variable of a sequence of i.i.d. random variables which represents the noisy data at the FC received from sensor $i$ through the noisy channel.

The FC estimates the correlation coefficient, required to calculate the mutual information for Chow-Liu algorithm in formula (\ref{eq:3}). The normalized estimator of the correlation coefficient for the noisy data is as follows:
\begin{equation}\label{eq:9}
\begin{aligned}
\begin{split}
\hat{\rho}_{ij} & = \frac{1}{n (1 + \sigma^2)} \Sigma_{k=1}^n Y_i^{(k)} Y_j^{(k)},\\
 & = \frac{1}{n (1 + \sigma^2)} \Sigma_{k=1}^n \bigg((X_i^{(k)} + N_i^{(k)})(X_j^{(k)} + N_j^{(k)})\bigg),\\
 & = \frac{1}{n (1 + \sigma^2)} \Sigma_{k=1}^n \bigg(X_i^{(k)} X_j^{(k)} + N_i^{(k)}N_j^{(k)} + X_i^{(k)}N_j^{(k)}+ X_j^{(k)}N_i^{(k)}\bigg).
\end{split}
\end{aligned}
\end{equation}

Incorrect ordering of mutual information event (also called the crossover event) is defined as follows \cite{Tan, F.Tan, Tavass}, \cite{Nikolakakis2}:
\begin{definition} [Crossover event/Incorrect Ordering]\label{def1}
Let $e$ be an edge in the original tree graph and let $e'$ be a pair of nodes $(r, s)$ such that the mutual information of these two edges has the following relation $I_e > I_{e'}$. The crossover event will occur if Chow-Liu algorithm estimates the mutual information in reverse order, i.e., $\hat{I}_e \leq \hat{I}_{e'}$.
\end{definition}

Using the definition of a crossover event and the positive correlation coefficient assumption, the probability of a crossover event can equivalently be defined as follows.
\begin{lem}[Equivalent definition] \label{lemma1} Let Y be the random variable which represents the data received at the FC from a machine, and is defined as $Y = X + N$ where $X \sim \mathcal{N}(0, 1)$ represents the original data while $N \sim \mathcal{N}(0,\sigma^2)$ is for Gaussian noise. The probability of a crossover event of a pair of edges $e = (i, j)$ and $e' = (r, s)$ with mutual information relation $I_e > I_{e'}$, is given in formula (\ref{eq:10}).
\begin{figure*}[!t]
\normalsize
\setcounter{mytempeqncnt}{\value{equation}}
\setcounter{equation}{9}
\begin{equation}\label{eq:10}
\begin{aligned}
Pr (\hat{I}_{e} \leq \hat{I}_{e'}) & \leq Pr\bigg(\frac{1}{n(1 + \sigma^2)}\sum_{k=1}^{n}(N_r^{(k)}N_s^{(k)} + X_r^{(k)}N_s^{(k)} + X_s^{(k)}N_r^{(k)} - N_i^{(k)}N_j^{(k)} - X_i^{(k)}N_j^{(k)} - \\
& X_j^{(k)}N_i^{(k)})\geq 0\bigg)+ Pr\bigg(\frac{1}{n(1 + \sigma^2)}\sum_{k=1}^{n}(X_r^{(k)}X_s^{(k)} - X_i^{(k)}X_j^{(k)}) \geq 0\bigg).
\end{aligned}
\end{equation}

\setcounter{equation}{\value{mytempeqncnt}}
\hrulefill
\vspace*{4pt}
\end{figure*}
\setcounter{equation}{10}
\end{lem}

\begin{proof}
See Appendix \ref{append1} for the proof.
\end{proof}
\begin{rem} \label{remark1}
  The right hand side of formula (\ref{eq:10}) has two parts; the second part (e.g., $Pr\bigg(\frac{1}{n(1 + \sigma^2)}\sum_{k=1}^{n}\\( X_r^{(k)}X_s^{(k)} - X_i^{(k)}X_j^{(k)}) \geq 0\bigg )$) corresponds to the probability of a crossover event (i.e., $Pr(\hat{\rho}_{e'} \geq \hat{\rho}_e)$) for finite size data set. For instance, consider the channels between sensors and the FC to be noiseless. If the crossover event happens, then it will occur due to finite size data set, and its probability is defined as $Pr\bigg(\frac{1}{n(1 + \sigma^2)}\sum_{k=1}^{n}(X_r^{(k)}X_s^{(k)} - X_i^{(k)}X_j^{(k)}) \geq 0\bigg )$.
\end{rem}

Next, the probability of a crossover event defined in Lemma \ref{lemma1} will be bounded from above. We call the following subsection the \textit{Probability of a Crossover Event in General Case} because different channels experience different noises with different means and variances in general. In addition, the condition in Lemma \ref{lemma1} is also relaxed with a small positive constant $t$.

\subsubsection{\textbf{Probability of a Crossover Event in General Case}}
It has been discussed that the communication  is distorted with different noises' strength in general. Let channels' noise $N_r^{(k)}$ and $N_i^{(k)}$ have the same distribution $\mathcal{N}(\mu_1,\sigma^2)$ while channels' noise  $N_s^{(k)}$ and $N_j^{(k)}$ also have the same distribution $\mathcal{N}(\mu_2,\sigma^2)$. For different channels noises, formula (\ref{eq:9}) can also be normalized by dividing with the bigger variance.

The strong condition in Lemma \ref{lemma1} is also relaxed by some small positive number $t$ (e.g., $Pr (\hat{I}_{e} + 2t \leq \hat{I}_{e'})$).  Lemma \ref{lemma1} has two parts as follows:
\begin{enumerate}
  \item Part 1: $Pr\bigg(\frac{1}{n(1 + \sigma^2)}\sum_{k=1}^{n}(N_r^{(k)}N_s^{(k)} + X_r^{(k)}N_s^{(k)} + X_s^{(k)}N_r^{(k)} - N_i^{(k)}N_j^{(k)} - X_i^{(k)}N_j^{(k)} - X_j^{(k)}N_i^{(k)})\geq t \bigg)$.
  \item Part 2: $Pr\bigg(\frac{1}{n(1 + \sigma^2)}\sum_{k=1}^{n}(X_r^{(k)}X_s^{(k)} - X_i^{(k)}X_j^{(k)}) \geq t \bigg)$.
\end{enumerate}

For the first part, let $L$ be the generic random variable of $L^{(k)} = O_1^{(k)}- O_2^{(k)}$, \textbf{for the difference of channel noise affection}, where $O_1^{(k)} = N_r^{(k)}N_s^{(k)} + X_r^{(k)}N_s^{(k)} + X_s^{(k)}N_r^{(k)}$ and $O_2^{(k)}= N_i^{(k)}N_j^{(k)} + X_i^{(k)}N_j^{(k)} + X_j^{(k)}N_i^{(k)}$ are related to the channel noise. Now the distribution of $L$ is needed.

The distribution of the product of two independent Gaussian random variables (e.g., $N_iN_j$) can be approximated by Normal distribution \cite{Macias} with parameters in equation (\ref{eq:6}) if the skewness approaches zero. Hence, the probability distribution of $L$ is  after some basic calculation using equation (\ref{eq:6}), is as follows: $\mathcal{N}(0,\sigma_L^2)$ where $\sigma_L^2 = 2\mu_2^2\sigma^2 + 2(\mu_1^2 + \sigma^2)\sigma^2 + 2\mu_1^2 + 2\mu_2^2 + 4\sigma^2$.

The following lemma bounds the probability of a crossover event for the general case.
\begin{lem}[The probability of a crossover event for the general case] \label{lemma2} Let $\{\textbf{y}^{(1)}, ..., \textbf{y}^{(n)}\}$ be $n$ independent samples received at the FC from the machines through Gaussian channels having different means and same variance where $\textbf{y}$ is a $d-$dimensional vector. Moreover, it is considered that the original data set $\{\textbf{x}^{(1)}, ..., \textbf{x}^{(n)}\}$ is approximately bounded data set such that $|x_i| \leq M\sigma$ for any component $i$ where $M\geq3$ and $\sigma$ is the standard deviation. Then the probability of a crossover event of a pair of edges $e = (i, j)$ and $e' = (r, s)$ is
\begin{enumerate}
  \item $Pr\bigg(\sum_{k = 1}^{n} L^{(k)} \geq tn(1 + \sigma^2)\bigg) \leq e^{-\frac{t^2 n (1 + \sigma^2)^2}{2 \sigma_L^2}}$ for Part 1;
  \item $Pr\bigg(\frac{1}{n} \sum_{k = 1}^{n}(Z^{(k)} - \mathbb{E}[Z^{(k)}]) \geq \alpha(1 + \sigma^2)\bigg) \leq e^{-\frac{2n\left[\alpha(1 + \sigma^2)\right]^2}{(b_M - a_M)^2}}$ for Part 2,
\end{enumerate}
where $\sigma^2$ is the variance of channel noise, the difference of channel noise affection ($L$) is distributed as $L\sim \mathcal{N}(0,\sigma_L^2)$ by our derivation above, and $\alpha = \frac{1}{1 + \sigma^2}\left[(\rho_{ij} - \rho_{rs}) + t\right]$. Moreover, $Z^{(k)} = X_r^{(k)}X_s^{(k)} - X_i^{(k)}X_j^{(k)}$ which is related to finite size data set without channel noise where $Z^{(k)} \in [a_M, b_M]$ for all $k$ and $-\infty < a_M \leq b_M < +\infty$.
\end{lem}

\begin{proof}
  Lemma \ref{lemma2} is proved in Appendix \ref{append2}.
\end{proof}

\begin{rem}\label{remark2}
  Gaussian distribution has unbounded support, however, it is well known that it has almost bounded support in the following sense: $Pr(|X - \mu| \leq 3\sigma) \simeq 0.997$ \cite{Bishop}. Therefore, the approximately bounded assumption about the original data set in Lemma \ref{lemma2} (e.g., $|x_i| \leq M\sigma$ for any $i$ where $M \geq 3$) is valid. In addition, the mutual information expression in formula (\ref{eq:3}) holds for the approximately bounded Gaussian random variables. Similarly, the correlation coefficient $\rho_{ij}$ and its estimator $\hat{\rho}_{ij}$ for Gaussian random variables are appropriate for estimating the correlation coefficient of approximately Gaussian random variables [Eq. 6 of \cite{Kugiumtzis}].
\end{rem}

\begin{rem}\label{remark3}
  The bound on the probability of a crossover event in Lemma \ref{lemma2}-1 is Chernoff bound for general case while Lemma \ref{lemma2}-2 is due to Hoeffding's inequality (\ref{eq:7}), and this bound (Lemma \ref{lemma2}) is exponentially decreasing for the number of samples.
\end{rem}

\subsubsection{\textbf{Probability of Incorrect Recovery}}
To bound the probability of incorrect recovery, it is assumed that the crossover event leads to incorrect tree structure recovery. The following theorem gives an upper bound on the probability of incorrect recovery using Theorem 1 of \cite{Tavass}.
\begin{thm}[Incorrect recovery probability for general case]\label{theorem1}
   Let $\{\textbf{y}^{(1)}, ..., \textbf{y}^{(n)}\}$ be $n$ samples where $\textbf{y}$ is a $d-$dimensional vector. The data set is received at the FC from all the machines through Gaussian channels with different means (e.g., $\mu_1$ and $\mu_2$) and same variance (e.g., $\sigma^2$). Moreover, the original data set $\{\textbf{x}^{(1)}, ..., \textbf{x}^{(n)}\}$ is approximately bounded data set such that $|x_i| \leq M\sigma$ for any $i$ where $M \geq 3$. Then the probability of incorrect tree structure recovery using this noisy data set is $$Pr(\hat{T}_p \neq T_p) \leq d^3 \left(e^{-\frac{t^2 n (1 + \sigma^2)^2}{2 \sigma_L^2}} + e^{-\frac{2n\left[\alpha(1 + \sigma^2)\right]^2}{(b_M - a_M)^2}}\right),$$
where $\alpha, \sigma^2, \sigma_L^2, a_M, b_M$, and $t$ are defined in Lemma \ref{lemma2}.
\end{thm}
\begin{proof} Using Theorem 1 of \cite{Tavass} (the prefactor $d^3$ is associated with all edges and all false edges) and Lemma \ref{lemma2}, the probability of incorrect recovery is
\begin{equation}\label{eq:11}
  \begin{split}
  Pr(\hat{T}_p \neq T_p) & \leq d^3 Pr(\hat{\rho}_e \leq \hat{\rho}_{e'}), \\
  & \leq d^3 \left( e^{-\frac{t^2 n (1 + \sigma^2)^2}{2 \sigma_L^2}} + e^{-\frac{2n\left[\alpha(1 + \sigma^2)\right]^2}{(b_M - a_M)^2}}\right).
  \end{split}
\end{equation}
\end{proof}

\begin{rem}\label{remark4}
  The prefactor $d^3$ is tight for chain structure tree, however, this prefactor can be reduced in many cases such as star structure which requires $d^2$ as the prefactor \cite{Tavass}. Moreover, the effect of this prefactor can be reduced by increasing more samples.

  It can be shown by setting the right hand side of formula (\ref{eq:11}) to $\delta$ and taking natural logarithm that Theorem \ref{theorem1} requires only $\mathcal{O}(\log(\frac{d}{\delta}))$ samples for the smallest sample size ($n$) to satisfy the failure probability upper bound $\delta > 0$ in recovering a $d$-node tree structure as compared to Theorem 3 of \cite{Nikolakakis} that needs $\mathcal{O}(\log^4(\frac{d}{\delta}))$ samples. Moreover, Theorem 3 of \cite{Nikolakakis} does not require the approximately bounded original data set $\{\textbf{x}^{(1)}, ..., \textbf{x}^{(n)}\}$.
\end{rem}

\subsection{Structure Learning in the Presence of Erasure Channels between Sensors and the Fusion Center (FC)}
In this subsection, the impact of Erasure channels on the tree structure recovery performance is studied, which exist between sensors and the FC. Due to an Erasure event, the data to the server may not be received. An Erasure event is also caused by straggler sensor nodes: sensor nodes that are significantly slower than the average \cite{Lee}, \cite{Buyukates}. We bound the probability of a crossover event and the probability of incorrect tree structure recovery in Lemma \ref{lemma3} and Theorem \ref{theorem2}, respectively.
\begin{figure}[t]
\centering
\includegraphics[width=2.5in]{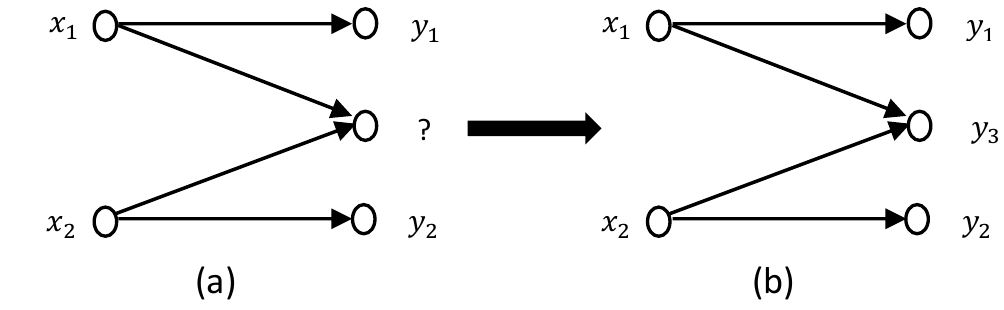}
\caption{Binary Erasure channel for a sensor before and after replacing the received symbol.}
\label{erasure}
\end{figure}

The random variable $X_i$ representing data at sensor $i$ is standard Gaussian, and the input alphabet is $\mathcal{X} \subseteq \mathcal{R}$. A sensor communicates the observed data to the FC through Erasure channel which is shown in Fig. \ref{erasure}a for binary alphabet, and the erased symbol $y_i = ?$ is replaced with \textbf{the symbol $y_3 = 1 \in \mathcal{Y}$ \cite{Yu} for estimating the correlation coefficient} as shown in Fig. \ref{erasure}b. This type of setting is used for packet loss problem which can be modeled by Erasure channel. The output alphabet $\mathcal{Y}$ is real as well, however, the random variable $Y_i$ representing the output is a mixed random variable.

The probability of Erasure event is defined as $\xi = Pr(Y = ? \not \in \mathcal{Y}) = Pr(Y = 1 \in \mathcal{Y})$ where $Y$ is the generic random variable for any output random variable with index $i$ and the sample index $k$, and the estimator of the correlation coefficient is $\hat{\rho}_e = \frac{1}{n} \sum_{k = 1}^{n} Y_i^{(k)}Y_j^{(k)}$. In addition, $Y = EX$ where $E$ is the random variable taking values in $\{1, ?\}$ representing an Erasure event and $X$ is the standard Gaussian random variable. The distribution of $Y$ is as follows:
\begin{equation}\label{eq:12}
Y =
  \begin{cases}
    X, & \mbox{when E = 1},\\
    1, & \mbox{when E = ?},
  \end{cases}
\end{equation}
then let $Pr(E = 1) = p$, we have $Pr(Y = 1) = 1 - p$, $Pr(Y > 1) = p\cdot Pr(X > 1)$, and $Pr(Y < 1) = p \cdot Pr(X < 1)$. The sum of these probabilities is equal to 1. \textbf{In addition, the probability of Erasure event $\xi = 1 - p$.}

\subsubsection{\textbf{Probability of a Crossover Event}} Using the positive correlation coefficient assumption, the probability of a crossover event is $Pr(\hat{I}_{e} \leq \hat{I}_{e'}) \equiv Pr (\hat{\rho}_e \leq \hat{\rho}_{e'})$ which is characterized here.
\begin{lem}[The probability of a crossover event for Erasure channel] \label{lemma3} Let the data set at the FC be $\{\textbf{y}^{(1)}, ..., \textbf{y}^{(n)}\}$ which is received through Erasure channels with the probability of an Erasure $\xi$. Furthermore, the original data set is approximately bounded Gaussian such that $|X_i| \leq M\sigma$ where $M \geq 3$, consequently, the received data set is bounded as well $|Y_i| \leq M\sigma$.

\noindent Then the probability of a crossover event of a pair of edges $e = (i, j)$ and $e' = (r, s)$ that follow the relation $I_e > I_{e'}$ is bounded from above by $$Pr(\hat{\rho}_e \leq \hat{\rho}_{e'}) \equiv Pr\left(\frac{1}{n} \sum_{k = 1}^{n}(Z^{(k)} - \mathbb{E}[Z^{(k)}]) \geq \beta\right) \leq e^{-\frac{2n\beta^2}{(b_M - a_M)^2}},$$ where $\beta = \rho_{e} - \rho_{e'}$ (difference of the correlation coefficients), $Z^{(k)} = Y_r^{(k)}Y_s^{(k)} - Y_i^{(k)}Y_j^{(k)}$, and $Z^{(k)} \in [a_M, b_M]$ for all $k$ and $-\infty < a_M \leq b_M < +\infty$. Moreover, $\rho_{e}$ and $\rho_{e'}$ are the correlation coefficients which are found by using the distribution of $Y$ while $\hat{\rho}_e$ and $\hat{\rho}_{e'}$ are the estimators of the correlation coefficients, which are unbiased.
\end{lem}

\begin{proof}
  The proof of this lemma is provided in Appendix \ref{append3}.
\end{proof}

\subsubsection{\textbf{Probability of Incorrect Recovery}} The crossover event causes the incorrect tree-structured GGM recovery, and this probability of incorrect recovery is derived in the following theorem using Theorem 1 of \cite{Tavass}. The probability of an Erasure event $\xi$ affects the probability of a crossover event in Lemma \ref{lemma3} implicitly, consequently, the probability of incorrect recovery in Theorem \ref{theorem2} below.

\begin{thm} [Incorrect recovery probability for Erasure channel] \label{theorem2} Let $n$ independent samples be given as $\{\textbf{x}^{(1)}, ..., \textbf{x}^{(n)}\}$, generated from a $d-$dimensional tree-structured  GGM such that the random variables have zero mean and unit variance and approximately bounded such that $|x_i| \leq M\sigma$ where $M \geq 3$. In addition, the received data set $\{\textbf{y}^{(1)}, ..., \textbf{y}^{(n)}\}$ is also bounded such that $|y_i| \leq M\sigma$. Having considered Erasure channels between sensors and the FC with the probability of an Erasure event $\xi$, the probability of incorrect tree structure recovery is $$Pr(\hat{T}_p \neq T_p) \leq  d^3 e^{-\frac{2n\beta^2}{(b_M - a_M)^2}},$$
where $\beta = \rho_{e} - \rho_{e'}$ (difference of the correlation coefficients), $Z^{(k)} = Y_r^{(k)}Y_s^{(k)} - Y_i^{(k)}Y_j^{(k)}$, and $Z^{(k)} \in [a_M, b_M]$.
\end{thm}

\begin{proof} Using the result from \cite{Tavass}, we have
\begin{equation}\label{eq:13}
  Pr(\hat{T}_p \neq T_p) \leq d^3 Pr (\hat{\rho}_e \leq \hat{\rho}_{e'}),
\end{equation}
where the prefactor $d^3$ is associated with all edges and all false edges.

Using Lemma \ref{lemma3}, the probability of incorrect recovery is
\begin{equation}\label{eq:14}
  Pr(\hat{T}_p \neq T_p) \leq d^3 e^{-\frac{2n\beta^2}{(b_M - a_M)^2}}.
\end{equation}
\end{proof}

\section{Structure Learning in Noisy Communication with Quantization}\label{sectionIV}
In this section, the sign method \cite{Tavass} is used to quantize the Gaussian variables into binary variables for communication efficiency, taking values in $\{-1, +1\}$. In the sign method, each sensor transmits the sign of the Gaussian random variables using the following source coding scheme: $U_i ^{(k)} = sign(X_i^{(k)})$ where $i$ and $k$ are sensor index and time index, respectively. 

In this section, it is assumed that there exist binary symmetric channels between sensors and the FC. The noise in these channels is time and sensor independent. Let $U_i$ and $\hat {U}_i$ represent the quantized data where $U_i$ represents the data at machine $i$ while $\hat {U}_i$ represents the data received at the FC through BSC. The received data is defined by $\hat{U}_i = R_i U_i$ where $\{R_i\}$ are i.i.d. Bernoulli random variables with probability $\epsilon$, taking values in the alphabet $\{-1, +1\}$.

The FC estimates the tree-structured GGM using the noisy quantized data set $\{\hat{\textbf{u}}^{(1)}, ..., \hat{\textbf{u}}^{(n)}\}$ where $\hat{\textbf{u}}^{(k)} \in \{-1, +1\}^d$. In addition, Tavassolipour \textit{et al.} \cite{Tavass} have shown that the sign method can preserve the true order of the mutual information. For the sign method, following equations are used to estimate the mutual information between two random variables $U_i$ and $U_j$:
\begin{equation}\label{eq:15}
  I(U_i ; U_j) = 1 - h(\theta_{ij}),
\end{equation}
where $h(\cdot)$ and $\theta_{ij}$ represent the binary entropy function and the probability related to correlation coefficient respectively, and are given by
\begin{equation}\label{eq:16}
  h(\theta_{ij}) = -\theta_{ij}\log(\theta_{ij}) - (1 - \theta_{ij})\log(1 - \theta_{ij}),
\end{equation}
\begin{equation}\label{eq:17}
  \theta_{ij} = \frac{1}{2} + \frac{arcsin(\rho_{ij})}{\pi}.
\end{equation}

The following estimator of $\theta_{ij}$ is optimal for the quantized data set in the sense that it is unbiased and has minimum variance \cite{Gamal},
\begin{equation}\label{eq:18}
  \hat {\theta}_{ij} = \frac{1}{n} \sum_{k=1}^{n} \mathcal{I}(U_i^{(k)} U_j^{(k)} = 1),
\end{equation}
where $\mathcal{I}(\cdot)$ is the indicator function. This estimated $\hat{\theta}_{ij}$ is used in equation (\ref{eq:15}) to calculate the mutual information, represented as $\hat{I}(U_i ; U_j)$.

\subsection{Probability of a Crossover Event} In this subsection, impact of communication noise is studied on the probability of crossover event for the noisy quantized data set in Lemma \ref{lemma4}.
\begin{lem}[The probability of a crossover event for the noisy quantized data set]\label{lemma4} Let $n$ independent noisy quantized samples be $\{\hat{\textbf{u}}^{(1)}, ..., \hat{\textbf{u}}^{(n)}\}$ where $\hat{\textbf{u}}^{(k)} \in \{-1, +1\}^d$, available to the FC having transmitted through $BSC(\epsilon)$. The nonquantized data set is generated from a $d-$dimensional tree-structured GGM $\mathcal{N(\bf 0, \Sigma)}$ such that $\mathbb{E}[X_i] = 1$ for any $i$. Then the probability of a crossover event due to channel noise of a pair of edges $e = (i, j)$ and $e' = (r, s)$ that follow the relation $\theta_e > \theta_{e'}$ (defined in (\ref{eq:17})), is given by
\begin{equation}\label{eq:19}
  Pr (\hat{\theta}_e  \leq \hat{\theta}_{e'}) \leq e^{nD},
\end{equation}
where $D = ln(p_0 + 2 \sqrt{p_1p_2}) \leq 0$ and
\begin{equation}\label{eq:20}
  p_0 = Pr(R_i R_j U_i U_j = R_r R_s U_r U_s),
\end{equation}
\begin{equation}\label{eq:21}
  p_1 = Pr(R_i R_j U_i U_j = -1, R_r R_s U_r U_s = 1),
\end{equation}
\begin{equation}\label{eq:22}
  p_2 = Pr(R_i R_j U_i U_j = 1, R_r R_s U_r U_s = -1),
\end{equation}
where $R_is$ are i.i.d. Bernoulli random variables taking values in the alphabet $\{-1, +1\}$ with probability $\epsilon$ and are independent with the data random variable $U_i$. Moreover, $\hat{\theta}_e$ is the estimator of $\theta_e$, which is defined in (\ref{eq:18}).
\end{lem}

\begin{proof}
  The proof of Lemma \ref{lemma4} is provided in Appendix \ref{append4}.
\end{proof}

\begin{rem}\label{remark5}
  The upper bound on the probability of a crossover event for the noisy quantized data set in formula (\ref{eq:19}) will increase, if the probability $\epsilon$ of Bernoulli random variable $R$ increases. In Lemma \ref{lemma4}, we study the impact of noisy communication and quantized data set due to the imperfect nature of the estimator of correlation coefficient.
\end{rem}

\subsection{Probability of Incorrect Recovery} The crossover event causes the incorrect tree-structured GGM recovery, and  the probability of incorrect recovery is bounded using Theorem 1 of \cite{Tavass}. Furthermore, the probabilities (\ref{eq:20} - \ref{eq:22}) in Lemma \ref{lemma4} do not have the closed-form solutions in general.
To derive the probability of incorrect recovery, Lemma \ref{lemma4} can be used similar to Theorem \ref{theorem1} and Theorem \ref{theorem2}.

\section{Algorithmic Bound}\label{sectionV}
In Theorem \ref{theorem1} and Theorem \ref{theorem2}, the prefactor $d^3$ (e.g., $d$-nodes tree structure) decreases the performances of the proposed bounds in comparison to the empirical performances especially for small data set. The bad effect of this prefactor can be reduced by increasing the sample size of the data set. However, it is very difficult to obtain the large data set sometimes using wireless sensor networks in the decentralized setting. Therefore, the Algorithmic Bound is proposed in this section to reduce the impact of this prefactor when some knowledge about the structure of the tree is known. This bound can be used for both Theorem \ref{theorem1} and Theorem \ref{theorem2} to tight their performances. To make it more specific, Algorithmic Bound is realized for Theorem \ref{theorem2} below.

\noindent \textbf{External Knowledge:} In this paper, we utilize the external knowledge about the tree-structured GGMs which is usually available in the form of the knowledge about the structure of the subtree of the original tree-structured GGMs and the upper bounds on the neighborhoods of nodes of the potential edges. The potential edge is defined to be the edge $\textbf{e}$ which connects any two subtrees to form an original tree or part of the original tree. For instance, in addition of the given data set $\{\textbf{y}^{(1)}, ..., \textbf{y}^{(n)}\}$ at the FC, we have some external knowledge about the tree-structured GGM $T_p(V, E)$ where $d = |V|$ as follow.
\begin{itemize}
  \item The tree-structured GGM $T_p(V,E)$ contains some disjoint subtrees with $d_1, d_2, ..., d_K$ nodes such that $d = d_1 + d_2 + ... + d_K$.
  \item Moreover, the upper bounds on the neighborhoods (or actual neighborhoods) of the potential edges are available.
\end{itemize}

Furthermore, it is better to have $d_1, ..., d_K < 10$ because of the impact of the prefactor $d^3$ in Theorem \ref{theorem2} especially for small data set. Suppose arrays $A_1$ and $A_2$ consist of disjoint subtrees information (e.g., $d_1, ..., d_K$) and the upper bounds on the neighborhoods (or actual neighborhoods) of the potential edges, respectively. Then Algorithmic Bound is given in Algorithm \ref{algorithm2} based on Theorem \ref{theorem2} for Erasure noise.

\begin{algorithm}[hbt]
\LinesNumbered
\SetAlgoLined
\SetKwProg{Fn}{Function}{:}{}
\KwIn{$A_1, A_2$ (these one dimensional arrays contains disjoint subtrees information and the upper bounds on the neighborhoods (or actual neighborhoods) of the potential edges respectively),
\\ \qquad \: $\beta = \rho_e - \rho_{e'}$ (difference of the correlation coefficients),
\\ \qquad \: $a_M, b_M$ (lower and upper bounds on the random variables in Theorem \ref{theorem2}),
\\ \qquad \: a number of samples $n$}
\KwOut{Upper bound on incorrect tree recovery probability $Pr(\hat{T}_p \neq T_p)$}
{
Initialization: set these arrays $U_1$, $U_2$ equal to zero\;
\tcp*[l]{compute the bound for each subtree}
\ForEach{$d_l \in A_1 $}{
    $u = d_l^3 e^{-\frac{2n\beta^2}{(b_M - a_M)^2}}$ \tcp*[r]{computing upper bounds for each subtree using Theorem \ref{theorem2} and $d_l = |V_l|$}
    Put the upper bound $u$ in array $U_1$\;
    }
\tcp*[r]{compute the bound for the neighborhoods of the potential edges}
\ForEach{$ N(q) \in A_2 $}{
    $v = N(q) e^{-\frac{2n\beta^2}{(b_M - a_M)^2}}$ \tcp*[r]{computing upper bound for a set of neighbors of node $q$ in array $A_2$}
    Put the upper bound $v$ in array $U_2$\;
    }

\tcp*[l]{sum all the parts of the bound}
Find the size of array $U_1$ \tcp*[r]{since $size(U_1) = size(U_2)$}

\For{$c = 1$ to $size(U_1)$}{
sum $\leftarrow$ sum + $U_1[c] + U_2[c]$ \tcp*[r]{sum of upper bounds}
}
\Return sum
}
\caption{Algorithmic Bound}
\label{algorithm2}
\end{algorithm}

\noindent \textbf{Description of Algorithmic Bound (e.g., Algorithm \ref{algorithm2}):} Algorithmic Bound is given in Algorithm \ref{algorithm2}. Given arrays $A_1, A_2$ that contains disjoint subtrees information and the upper bounds on the neighborhoods (or the actual neighborhoods), respectively, and the parameters of Theorem \ref{theorem2}, Algorithm \ref{algorithm2} calculates the upper bound for each subtree in array $A_1$ (e.g., line 3-6), then it calculates the upper bound for each neighborhoods of the potential edges given in array $A_2$ (e.g., line 8-11), and finally it adds all the upper bounds calculated so far in line 14-16.

To prove the validity of Algorithm \ref{algorithm2}, we use the dominant crossover event in tree-structured GGMs which is defined in \cite{F.Tan}, \cite{Tavass}. The effect of the dominant crossover event is as follows.

\begin{definition}[Impact of dominant errors]\label{def2}
An edge $e_{ij}$ is replaced with one of the candidate false edges: node $i$ is connected to one of the neighbors of node $j$ or vice versa.
\end{definition}

\noindent \textbf{Analysis of Algorithm \ref{algorithm2}:} First, the time complexity of Algorithmic Bound (Algorithm \ref{algorithm2}) is discussed, and then the proof of Algorithmic Bound correctness is provided by showing the validity of upper bound which is tighter than Theorem \ref{theorem2} because it reduces the effect of prefactor $d^3$ for tree-structured recovery.

\noindent \textbf{Time Complexity:} The overall time complexity of Algorithmic Bound (Algorithm \ref{algorithm2}) is $\mathcal{O}(m)$ where $m$ represents the size of array $A_1$ or $A_2$.

\begin{thm}[Algorithmic Bound on incorrect recovery probability] \label{theorem3}
Algorithmic Bound (Algorithm \ref{algorithm2}) gives the valid upper bound on the incorrect tree structure recovery probability $Pr(\hat{T}_p \neq T_p)$, and it has $\mathcal{O}(m)$ time complexity where $m$ is the number of subtrees.
\end{thm}
\begin{proof}
  Please see Appendix \ref{append5} for the proof.
\end{proof}

\begin{figure}[ht]
\centering
\begin{subfigure}{.4\textwidth}
  \centering
  \includegraphics[width=1.5in]{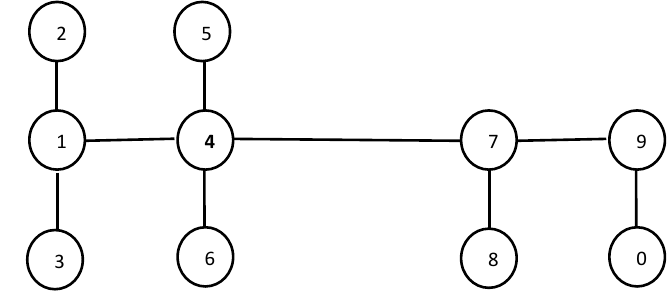}
  \caption{A complete tree.}
  \label{wholeT}
\end{subfigure}
\begin{subfigure}{.4\textwidth}
  \centering
  \includegraphics[width=1.5in]{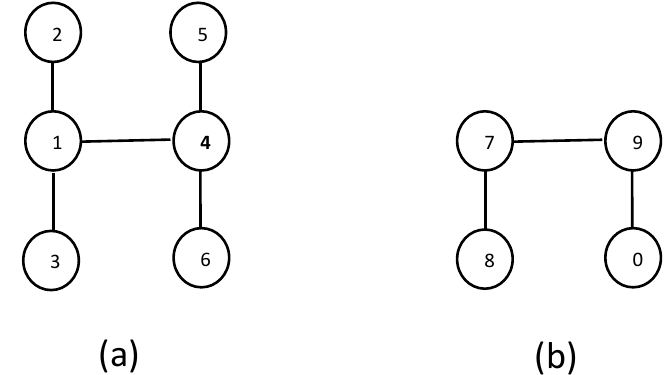}
  \caption{Fig. \ref{wholeT} is broken into two subtrees with 6 and 4 nodes.}
  \label{subtreesT}
\end{subfigure}
\label{subtree}
\begin{subfigure}{.4\textwidth}
  \centering
  \includegraphics[width=1.5in]{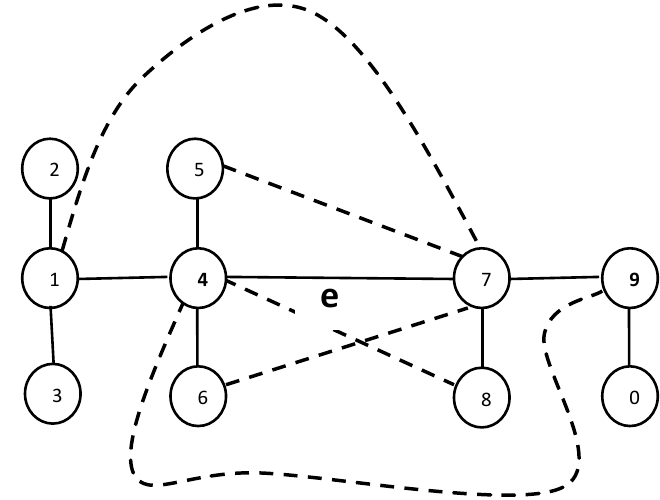}
  \caption{Possible false candidate edges $\{(1,7), (4,8), (4,9),$ $ (5,7), (6,7)\}$ in replacement of the original edge $e$.}
\label{possibleE}
\end{subfigure}
\caption{Illustration of Algorithmic Bound (Algorithm \ref{algorithm2}).}
\label{illustration}
\end{figure}

\noindent \textbf{Example 1 (Illustration of Algorithmic Bound):} A complete tree is given in Fig. \ref{wholeT} from which the data set is generated. This is the original tree which we want to estimate from the noisy data set. The external knowledge about this original tree (Fig. \ref{wholeT}) is the knowledge about numbers of nodes in each subtrees of the original one. The numbers of nodes in each subtrees are 6 and 4 such as shown in Fig. \ref{subtreesT}. Moreover, the neighborhoods of node 4 and node 7 are given as the external knowledge. These neighborhoods are $N(4) = \{1, 5, 6\}$ and $N(7) = \{8, 9\}$. We require the neighborhoods of the nodes of the potential edge $(4, 7)$ because these are the nodes for false candidate edges due to the dominant crossover event (e.g., Definition \ref{def2}) as shown in Fig. \ref{possibleE}.

\noindent In comparison to Theorem \ref{algorithm2}, the upper bound, provided by Algorithmic Bound, will be much tighter for fix sample size $n$ due to the reduction of prefactor $d^3$ in formula (\ref{eq:14}).

\section{Experiments}\label{sectionVI}
In this section, our results for Gaussian, Erasure channels, and BSC are simulated where Gaussian channels are considered for the general case.  In the general case, different sensors' communication is affected by different Gaussian noise. Next, the data set generation method is given, and then our results for Theorem \ref{theorem1}, Theorem \ref{theorem2}, Algorithmic Bound, and Lemma \ref{lemma4} for quantized data set are shown.

We use the error probability metric for all the simulations, which is defined as
\begin{equation}\label{eq:23}
  Error\; Probability = \frac{\#\; of incorrect\; estimated\; trees}{\#\; of algorithm\; runs}.
\end{equation}

\subsection{Synthetic Data}
In order to simulate Theorem \ref{theorem1}, Theorem \ref{theorem2}, and Algorithmic Bound, the synthetic data set is generated from a random tree with $d$ nodes as follows.
\begin{enumerate}
  \item A random tree with $d$ nodes is generated, and the random weight is assigned to each edge of the tree corresponding to the correlation coefficient between two nodes of the edge (i.e., the random weight is set for each edge from this interval $[0.1$, $0.9]$).
  \item The correlation coefficients for non-neighboring nodes (i.e., $x_i$ and $x_m$) can be computed as the product of the correlations on the shortest path from $x_i$ to $x_m$.
  \item The covariance matrix of the tree-structured GGM is generated using the weighted tree.
  \item Finally, $n$ independent samples are generated from zero mean multivariate (standard) normal distribution with covariance matrix (in the previous step), and distributed among $d$ sensors vertically.
\end{enumerate}

\subsection{Impact of Gaussian noise on the Empirical Performance}\label{cases}
\begin{figure}
\centering
\includegraphics[width=0.8in]{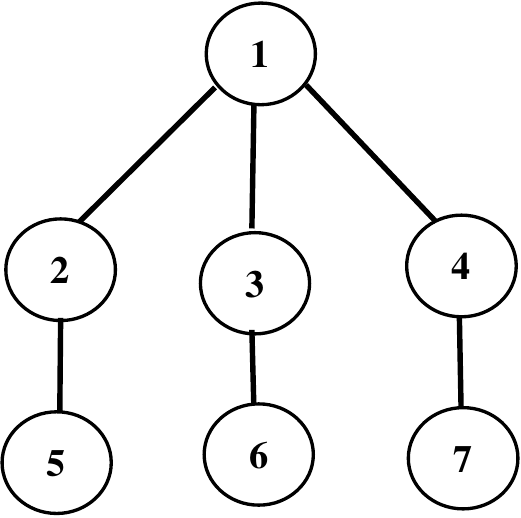}
\caption{Example of tree-structured graph.}
\label{Fig4}
\end{figure}

In this subsection, the impact of non-identically distributed noise on the empirical performance of tree-structured recovery is studied for Fig. \ref{Fig4}.

The data set is generated from the multivariate Gaussian distribution with tree-structured model in Fig. \ref{Fig4} using the method discussed previously with the edge weight in the interval $[0.1, 0.9]$. The range of the data set is fixed in the interval $[-3, 3]$. For the simulation, two cases are considered as follows.
\begin{itemize}
  \item \textit{Case 1:} In this case, the data sets from nodes 1, 5, 6, and 7 are corrupted by noisy random variable from Gaussian distribution $\mathcal{N}(1, 1)$ while the data sets of remaining nodes (e.g., 2, 3, and 4) are made noisy by Gaussian noise from $\mathcal{N}(0.05, 1)$.
  \item \textit{Case 2:} In this case, the data sets from nodes 1, 5, 6, and 7 are corrupted by noisy random variable from Gaussian distribution $\mathcal{N}(1, 2)$ while the data sets of remaining nodes (e.g., 2, 3, and 4) are made noisy by Gaussian noise from $\mathcal{N}(0.05, 2)$.
\end{itemize}

\begin{figure}
\centering
\includegraphics[width=2.5in]{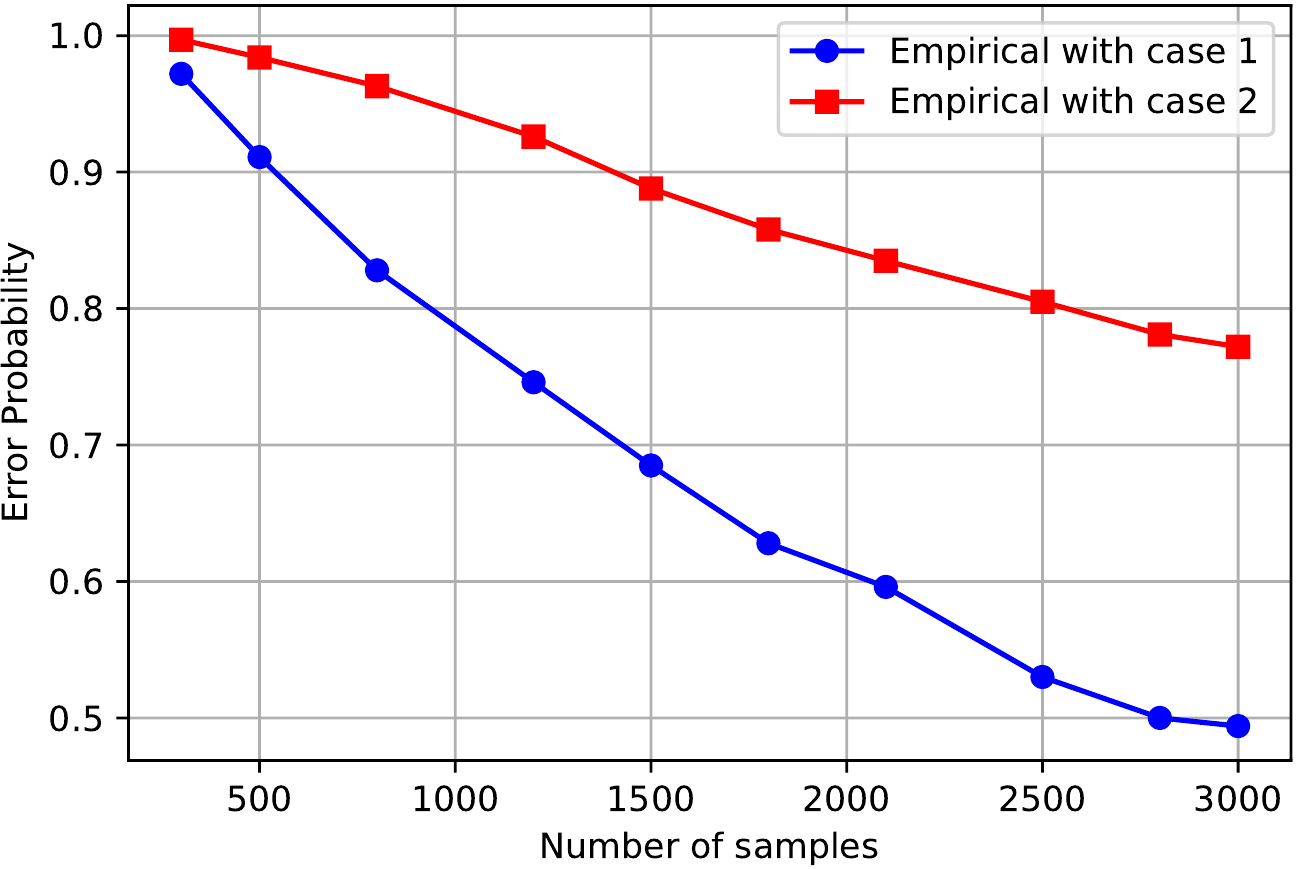}
\caption{Example of tree-structured graph.}
\label{noiseImpact}
\end{figure}

The noise impact result is shown in Fig. \ref{noiseImpact} where increasing the variance of noises in \textit{Case 2} strongly deteriorates the tree-structured GGM recovery performance. For instance, structure recovery error probability of \textit{Case 1} reaches 0.5 value for about 3000 samples while for \textit{Case 2}, structure recovery probability is about 0.8 for the same number of samples.

\subsection{Simulation result for Theorem \ref{theorem1}}
We consider the tree-structured graph in Fig. \ref{Fig4} for comparing the empirical performance and the result of Theorem \ref{theorem1}. For performance comparison, we use all the settings of \textit{Case 1} in the previous subsection where the data sets from nodes 1, 5, 6, and 7 are corrupted by noisy random variable from Gaussian distribution $\mathcal{N}(1, 1)$ while the data sets of remaining nodes (e.g., 2, 3, and 4) are made noisy by Gaussian noise from $\mathcal{N}(0.05, 1)$.

\begin{figure}
\centering
\includegraphics[width=2.5in]{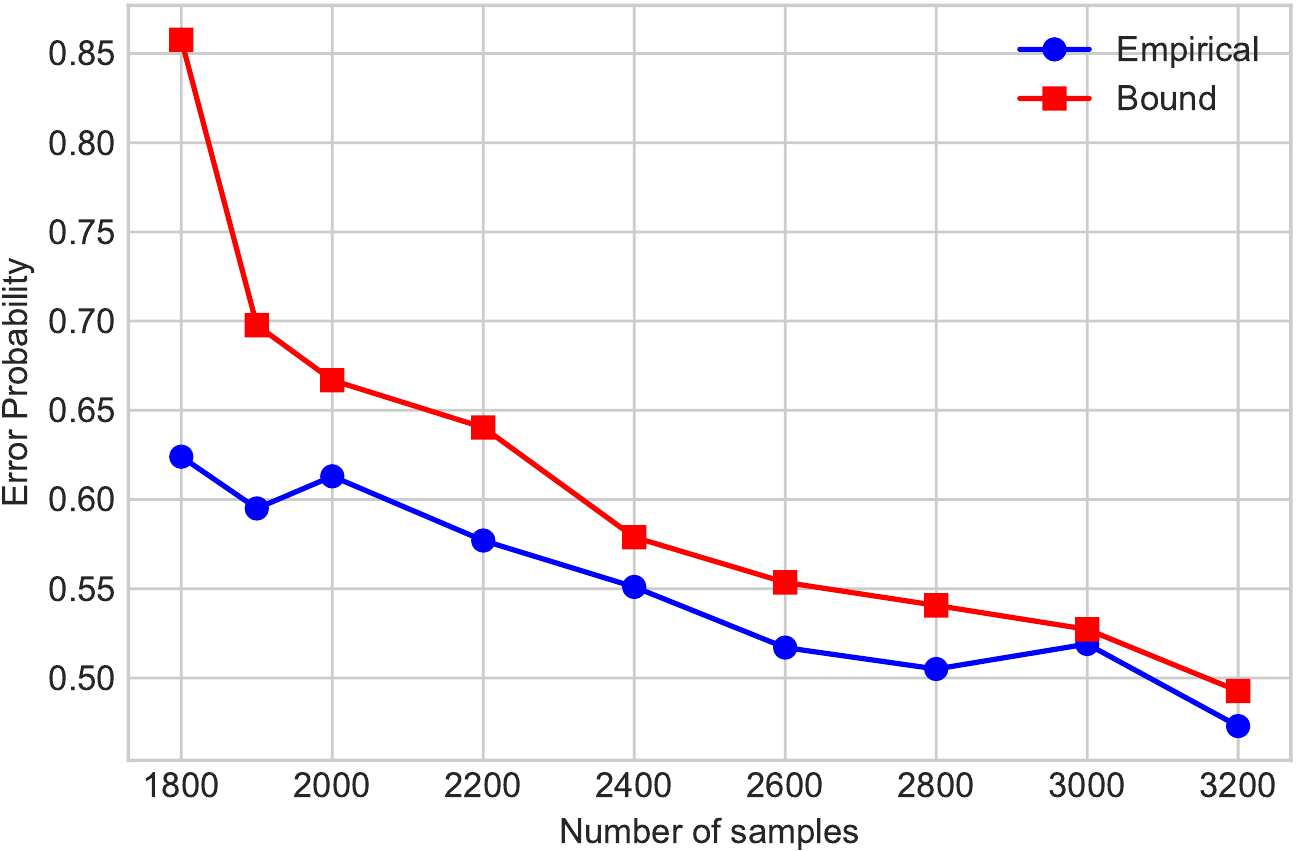}
\caption{Empirical error probability and Theorem \ref{theorem1} bound for Fig. \ref{Fig4}.}
\label{Fig5}
\end{figure}

The result is shown in Fig. \ref{Fig5} where it can be observed that error probability decreases significantly as the number of samples increases. A significant gap exists between empirical and Theorem \ref{theorem1} performances for sample size between 1800 - 2400 as shown in Fig. \ref{Fig5}, which is due to the small positive constant $t$ in Theorem \ref{theorem1}. In this experiment, the range of $t$ is set to be in the interval $[0.1, 0.14]$. As the sample size increases (between 2400-3200), the performance becomes tight. Moreover, the performance can be optimized further by changing the range of $t$. The error probability is calculated using formula (\ref{eq:23}) by running the experiment 1000 times.

\subsection{Empirical Comparison of Sample Complexity in Theorem \ref{theorem1} and Theorem 3 of \cite{Nikolakakis}}
We have compared our theoretical result of Theorem \ref{theorem1} with Theorem 3 of \cite{Nikolakakis} in Remark \ref{remark4}. For empirical comparison of sample complexity ($n$) of our Theorem \ref{theorem1} with Theorem 3 of \cite{Nikolakakis}, Fig. \ref{Fig4} is used with \textit{Case 1} in \ref{cases}.

\begin{table*} [!t]
\caption{Sample Complexity Comparison of Theorem \ref{theorem1} with Theorem 3 in \cite{Nikolakakis}.}
\label {table:1}
\centering
\scalebox{0.7}{
\begin{tabular}{ | c | c | c | }
\hline
Upper bound on $Pr(\hat{T}_p \not = T)$ & Number of samples $(n) \times 10^3$ (in Theorem \ref{theorem1}) & Number of samples $(n) \times 10^6$ (in Theorem 3 of \cite{Nikolakakis}) \\ [0.5 ex]
 \hline
0.9 & $1.18$ & $11.5$  \\
0.8 & $1.21$ & $12.2$  \\
0.7 & $1.23$ & $13.03$  \\
0.6 & $1.26$ & $14.03$  \\
0.5 & $1.3$ & $15.3$ \\ 
0.4 & $1.34$ & $16.97$  \\
0.3 & $1.4$ & $19.31$  \\
0.2 & $1.48$ & $23.01$  \\
0.1 & $1.62$ & $30.55$  \\
\hline
\end{tabular}}
\end{table*}

The parameters $R = 0.1$ of Theorem 3 \cite{Nikolakakis} and $t = 0.1$ of our Theorem \ref{theorem1} in formula (\ref{eq:11}) are set. The numerical results are shown in Table \ref{table:1} where it can be observed that the proposed upper bound in Theorem \ref{theorem1} for hidden Gaussian model with different noise intensities has better performance as compared to Theorem 3 \cite{Nikolakakis}. The reason for huge performance gap is due to the approximately bounded Gaussian tree-structured model and the positive correlation coefficient assumption, and the proposed Theorem \ref{theorem1} can only be utilized for specific applications.

\subsection{Evaluation of Theorem \ref{theorem2} and Algorithmic Bound}
In this subsection, Algorithmic Bound performance is compared with Theorem \ref{theorem2} for the tree graph in Fig. \ref{wholeT}.

\begin{figure}[t]
\centering
\includegraphics[width=2.5in]{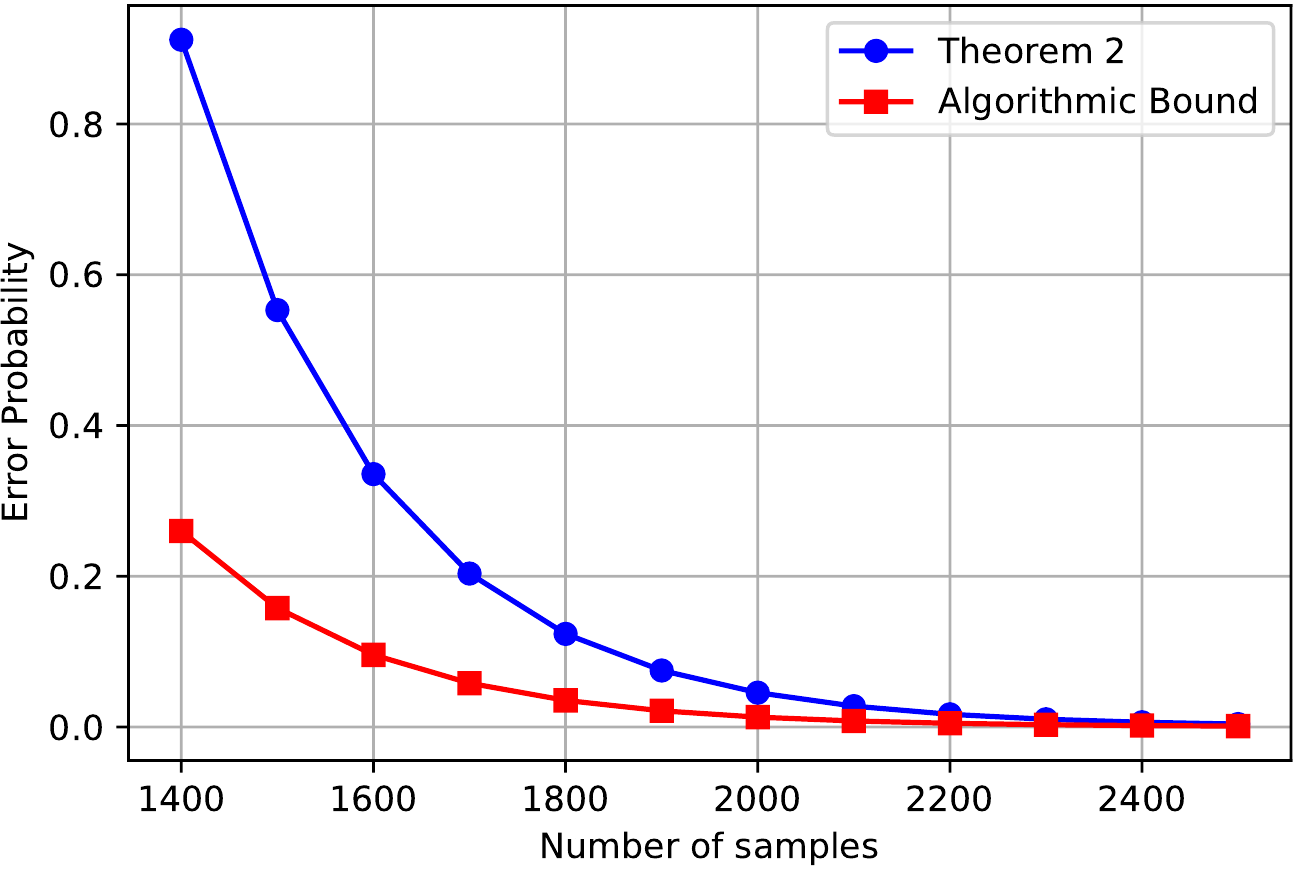}
\caption{Theorem \ref{theorem2} and Algorithmic Bound for Fig. \ref{wholeT}.}
\label{algorithmic}
\end{figure}

Fig. \ref{algorithmic} shows the comparison of Theorem \ref{theorem2} and Algorithmic Bound performances for the tree graph in Fig. \ref{wholeT}. It can be observed that the Algorithmic Bound is very tight compared to Theorem \ref{theorem2} especially when the number of samples is between $[1400, 1800]$. That is the objective for proposing Algorithmic Bound (Algorithm \ref{algorithm2}). As the number of samples increases further, the difference in performances of Theorem \ref{theorem2} and Algorithmic Bound gets insignificant.

\subsection{Evaluation of Lemma 4}
We discussed that the probabilities $p_0$, $p_1$, and $p_2$ (e.g., formulas (\ref{eq:20}-\ref{eq:22})) do not have closed-form solutions in general. However, if the two edges $e$ and $e'$ share a common node, these probabilities can be evaluated analytically (i.e., equations (18-20) in \cite{Tavass}).

\begin{figure}[t]
\centering
\includegraphics[width=1in]{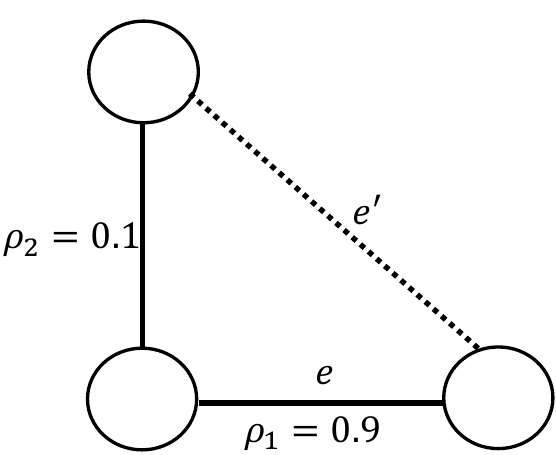}
\caption{A sample subtree of three nodes for evaluation of Lemma \ref{lemma4}.}
\label{model_4}
\end{figure}

\begin{figure}[ht]
\centering
\includegraphics[width=2.5in]{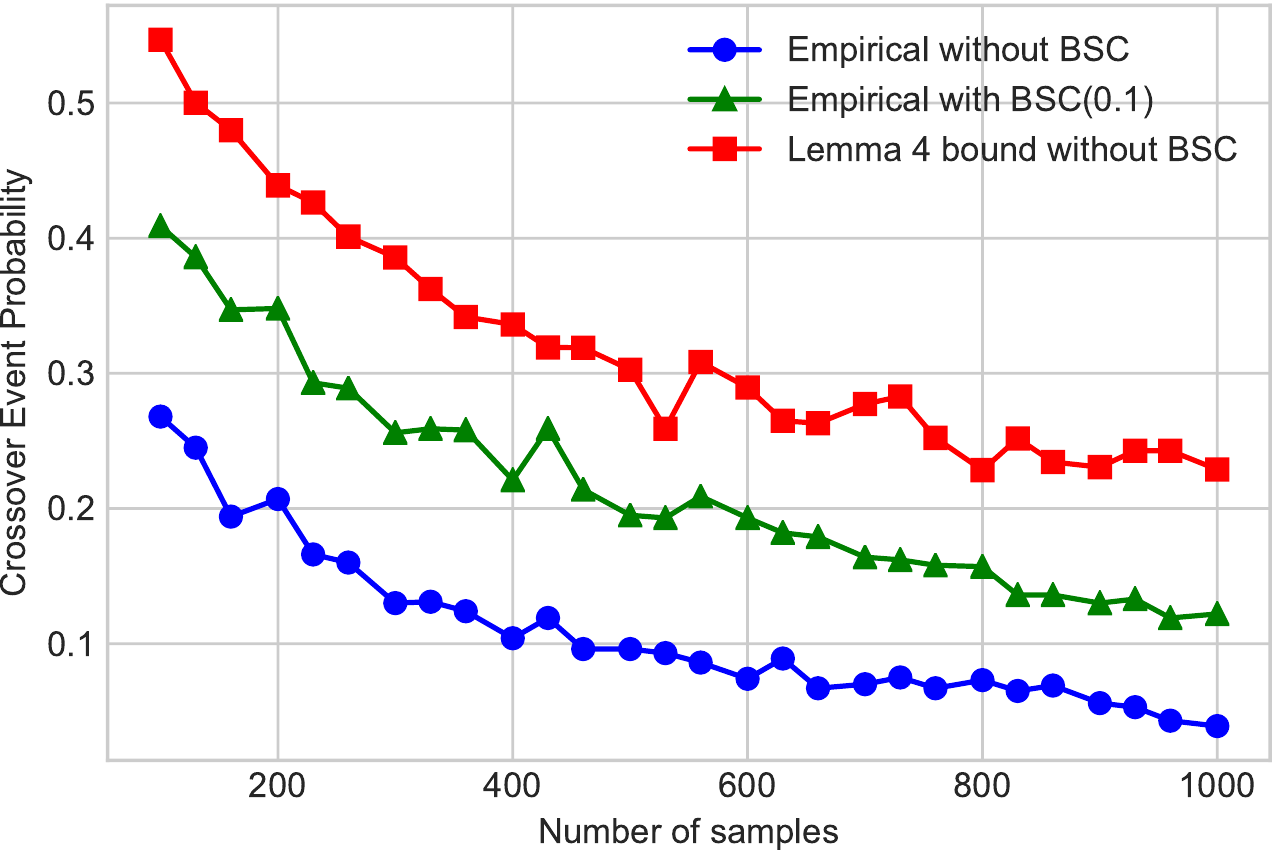}
\caption{Probability of a crossover event for $e$ and $e'$ in Fig. \ref{model_4} with BSC error probability $\epsilon = 0.1$.}
\label{epsilon3}
\end{figure}

The data set for Fig. \ref{model_4} \cite{Tavass} is generated with the correlation coefficients
$\rho_1 = 0.9$ and $\rho_2 = 0.1$ of the edges. The results are shown in Fig. \ref{epsilon3} where the BSC error probability  is 10\% (i.e., $\epsilon = 0.1$).

The performance of Lemma \ref{lemma4} is compared with empirical probabilities of crossover events with and without BSC channels between sensors and the FC, which are calculated using formula (\ref{eq:23}) by running the algorithm 1000 times. A crossover event will happen if Chow-Liu algorithm selects the edge $e'$ in replacement of the actual edge $e$ as shown in Fig. \ref{model_4}. Fig. \ref{epsilon3} shows that Lemma \ref{lemma4} bound performance is loose as compared to empirical performances because of considering two types of noise: channel noise and finite size data set.

\subsection{Learning Star-Structured Tree}
\begin{figure}[t]
\centering
\includegraphics[width=2.5in]{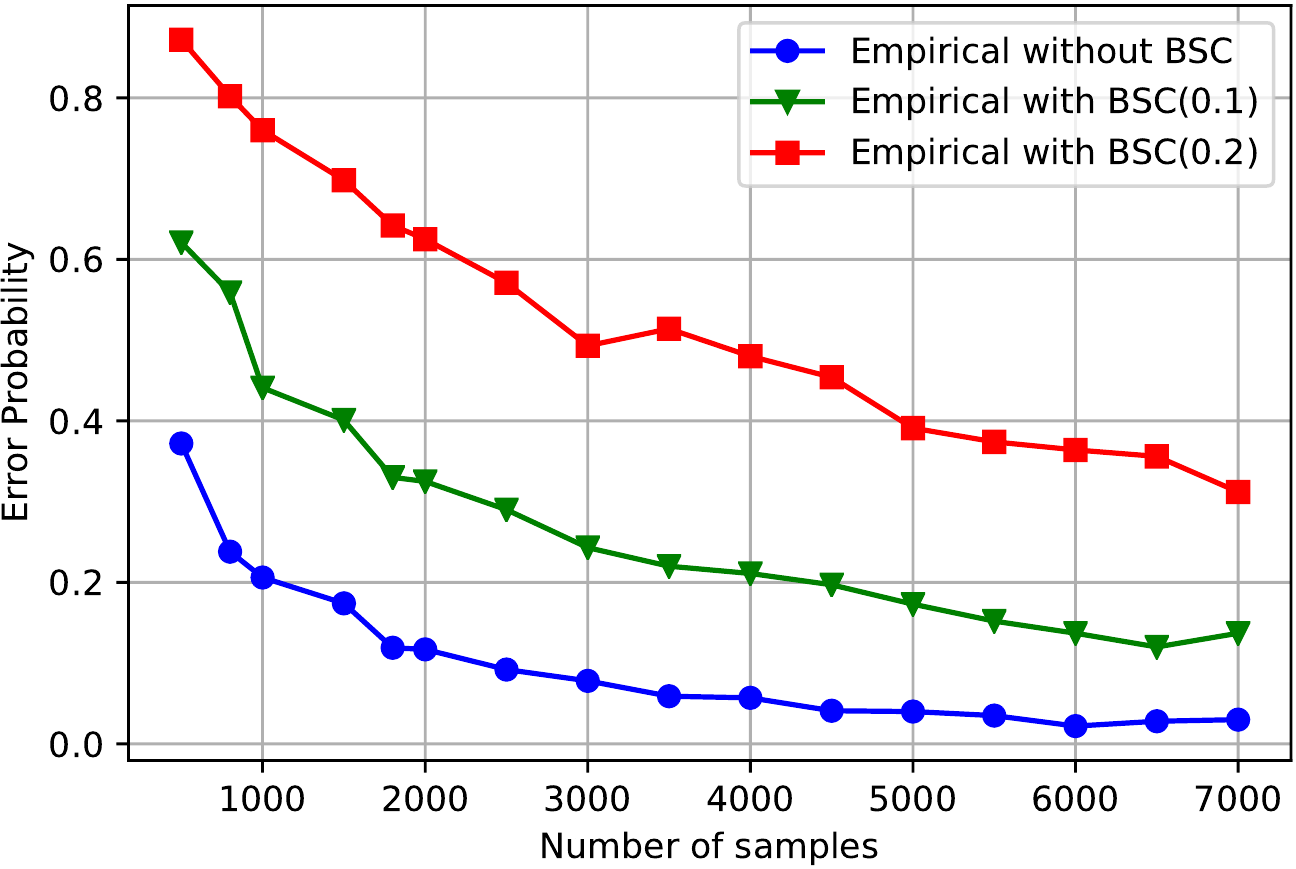}
\caption{Empirical error probability for the star-structured tree with $d = 5$, $\epsilon = 0.1$, and $\epsilon = 0.2$.}
\label{star}
\end{figure}

It has been observed that the star-structured tree model is the most difficult to learn for fixed parameters \cite{F.Tan}. Fig. \ref{star} shows the empirical performance of Chow-Liu algorithm for five nodes star-structured tree with quantized data set where the weight of the edge is chosen from the range $[0.1, 0.9]$ randomly. It can be seen that more samples are required for learning the star-structured tree model with five nodes. The empirical performance with and without error has been plotted
in Fig. \ref{star} where it can be observed that empirical performance with BSC($0.1$) requires about 4000 samples to achieve error probability $0.2$ as compared to empirical performance without error which achieves this performance with about 1000 samples.

\section{Conclusion}\label{sectionVII}
In this paper, the impact of noisy channels is studied between sensors and the FC for decentralized learning of tree-structured GGMs with finite size data set. The proposed decentralized learning uses the Chow-Liu algorithm for estimating the tree-structured GGM. Three types of noisy channels: Gaussian, Erasure, and BSC are considered, and the upper bounds on the probability of incorrect tree structure recovery were derived. Besides, an algorithm has been developed to tight the performances of Theorem \ref{theorem1} or Theorem \ref{theorem2} especially for small data set when some external knowledge is available. We validated the proposed upper bounds for these channels between sensors and the FC using the synthetic data sets. 

\appendices

\section{Proof of Lemma \ref{lemma1}}\label{append1}
\begin{proof}
  Since the mutual information is an increasing function of the squared correlation coefficient, shown in equations (\ref{eq:3}-\ref{eq:4}), then the estimator of mutual information is related to the estimator of the correlation coefficient as
\begin{equation}\label{eq:24}
  Pr(\hat{I}_{e} \leq \hat{I}_{e'}) \equiv Pr(\hat{\rho}_e^2 \leq \hat{\rho}_{e'}^2).
\end{equation}

Using the positive correlation coefficient assumption, the probability of a crossover event is equivalently defined as
\begin{equation}\label{eq:25}
  Pr(\hat{\rho}_e^2 \leq \hat{\rho}_{e'}^2) \equiv Pr(\hat{\rho}_e \leq \hat{\rho}_{e'}).
\end{equation}

The difference of the estimators of the correlation coefficients and the probability of a crossover event are as follows using equation (\ref{eq:9}).
\begin{equation}\label{eq:26}
\begin{aligned}
\hat{\rho}_{e'} - \hat{\rho}_{e} & = \frac{1}{n(1 + \sigma^2)}\sum_{k=1}^{n}\bigg( X_r^{(k)}X_s^{(k)} + N_r^{(k)}N_s^{(k)} + X_r^{(k)}N_s^{(k)} + X_s^{(k)}N_r^{(k)} -
X_i^{(k)}X_j^{(k)} - N_i^{(k)}N_j^{(k)}\\
& - X_i^{(k)}N_j^{(k)} - X_j^{(k)}N_i^{(k)}\bigg),
\end{aligned}
\end{equation}
\begin{equation}\label{eq:27}
\begin{aligned}
  Pr(\hat{\rho}_e \leq \hat{\rho}_{e'}) & = Pr\bigg(\frac{1}{n(1 + \sigma^2)}\sum_{k=1}^{n}(X_r^{(k)}X_s^{(k)} + N_r^{(k)}N_s^{(k)} + X_r^{(k)}N_s^{(k)} +
  X_s^{(k)}N_r^{(k)} - X_i^{(k)}X_j^{(k)}\\
  & - N_i^{(k)}N_j^{(k)} - X_i^{(k)}N_j^{(k)} - X_j^{(k)}N_i^{(k)}) \geq 0 \bigg).
\end{aligned}
\end{equation}

Let $A = \{X \geq 0\}$, $B = \{Y \geq 0\}$, and $C = \{X + Y \geq 0\}$ be the events such that $X = \frac{1}{n(1 + \sigma^2)}\sum_{k=1}^{n}\bigg( N_r^{(k)}N_s^{(k)} + X_r^{(k)}N_s^{(k)} + X_s^{(k)}N_r^{(k)} - N_i^{(k)}N_j^{(k)} - X_i^{(k)}N_j^{(k)} - X_j^{(k)}N_i^{(k)} \bigg)$ and \\$Y = \frac{1}{n(1 + \sigma^2)}\sum_{k=1}^{n}\bigg( X_r^{(k)}X_s^{(k)} - X_i^{(k)}X_j^{(k)}\bigg)$ are the random variables, then
\begin{equation}\label{eq:28}
\begin{split}
Pr(\hat{\rho}_{e'} - \hat{\rho}_{e}\geq 0) = Pr(C) & \stackrel{(a)}{\leq} Pr(A \cup B),\\
& \stackrel{(b)}{\leq} Pr(A) + Pr(B),
\end{split}
\end{equation}
where $(a)$ can be proved by showing that $C \subseteq A \cup B$ using event definitions while $(b)$ is due to sum rule of probability.
\end{proof}

\section{Proof of Lemma \ref{lemma2}}\label{append2}
\begin{proof}
\textbf{Part 1:} The difference of channel noise affection ($L$) is distributed according to $L\sim \mathcal{N}(0,\sigma_L^2)$. The probability of a crossover event can be obtained as
\begin{equation}\label{eq:29}
\begin{split}
Pr\left(\sum_{k = 1}^{n} L^{(k)} \geq tn(1 + \sigma^2)\right) & = Pr\left( e^{\lambda \sum_{k = 1}^{n} L^{(k)}} \geq e^{\lambda tn(1 + \sigma^2)}\right), \quad \lambda > 0\\
  & \stackrel{(a)}{\leq} \frac{\mathbb{E}\left[e^{\lambda \sum_{k=1}^{n} L^{(k)}}\right]}{e^{\lambda tn(1 + \sigma^2)}}, \\
  & \stackrel{(b)}{=} \frac{\left(\mathbb{E}\left[e^{\lambda L}\right]\right )^n}{e^{\lambda tn(1 + \sigma^2)}},
\end{split}
\end{equation}
where $L$ is the generic random variable of $L^{(k)}$, and $(a)$ and $(b)$ are due to Markov's inequality and time independent difference of channel noise affection, respectively.

By optimizing $\lambda$ and using the moment generating function of $L \sim \mathcal{N}(0, \sigma_L^2)$, the probability of a crossover event is upper bounded by
\begin{equation}\label{eq:30}
  Pr\left(\sum_{k = 1}^{n} L^{(k)} \geq tn(1 + \sigma^2)\right) \leq e^{\frac{-t^2 n (1 + \sigma^2)^2}{2 \sigma_L^2}},
\end{equation}
where $\sigma_L^2$ is the variance of $L$.

\noindent \textbf{Part 2:} It is known that all the $X_i$ are Gaussian random variables with zero mean and unit variance with approximately bounded assumption $|X| \leq M\sigma$ where $M \geq 3$ where $\sigma$ is the standard deviation. Let $\varrho_{ij}$ and $\hat{\varrho}_{ij}$ be the correlation coefficient and the estimator of the correlation coefficient for the approximately bounded Gaussian random variables.

Furthermore, the relation between the correlation coefficients of approximately bounded Gaussian random variables $\varrho_{ij}$ and Gaussian random variables $\rho_{ij}$ is $\varrho_{ij} \approx - \rho_{ij}$ [Eq. 6 of \cite{Kugiumtzis}]. This approximation is tight for the approximately bounded Gaussian random variables with the range $|X| \leq M\sigma$ where $M \geq 3$. Hence, we can utilize the estimator of correlation coefficient $\hat{\rho}_{ij}$ for Gaussian case for estimating the correlation coefficient of approximately bounded Gaussian random variables. In addition, the negative sign has no effect for tree structure learning because one needs the magnitude of the correlation coefficient for estimating the mutual information that is related to squared of the correlation coefficient (\ref{eq:3}).

The estimator of the correlation coefficient is unbiased such that
\begin{equation}\label{eq:31}
  \mathbb{E}[\hat{\rho}_{ij}] =\mathbb{E}[\frac{1}{n}\sum_{k = 1}^{n}X_i^{(k)}X_j^{(k)}] = \rho_{ij}
\end{equation}
for any $(i , j)$, where $\hat{\rho}_{ij}$ is a random variable but $\rho_{ij}$ is not. Moreover, $\rho_{ij} = \mathbb{E}[X_i X_j]$ is the correlation coefficient of standard Gaussian random variables $X_i$ and $X_j$.

By using equation (\ref{eq:31}), the crossover event $\hat{\rho}_{rs} - \hat{\rho}_{ij} \geq t$ is equivalent to
\begin{equation}\label{eq:32}
\begin{split}
   \hat{\rho}_{rs} - \hat{\rho}_{ij} - \mathbb{E}[\hat{\rho}_{rs} - \hat{\rho}_{ij}] & \geq \rho_{ij} - \rho_{rs} + t, \\
     \frac{1}{1 + \sigma^2} (\hat{\rho}_{rs} - \hat{\rho}_{ij} - \mathbb{E}[\hat{\rho}_{rs} - \hat{\rho}_{ij}]) & \geq \frac{1}{1 + \sigma^2}(\rho_{ij} - \rho_{rs} + t),\\
     \frac{1}{n(1 + \sigma^2)} \sum_{k = 1}^{n}((X_r^{(k)}X_s^{(k)} - X_i^{(k)}X_j^{(k)}) - \mathbb{E}[X_r^{(k)}X_s^{(k)} - X_i^{(k)}X_j^{(k)}]) & \geq \frac{1}{1 + \sigma^2}(\rho_{ij} - \rho_{rs} + t).
\end{split}
\end{equation}

Let $\alpha = \frac{1}{1 + \sigma^2}(\rho_{ij} - \rho_{rs} + t)$ and $Z^{(k)} = X_r^{(k)}X_s^{(k)} - X_i^{(k)}X_j^{(k)}$ which is related to the crossover event due to finite size data set. Then formula (\ref{eq:32}) becomes
\begin{equation}\label{eq:33}
\begin{split}
  \frac{1}{n(1 + \sigma^2)} \sum_{k = 1}^{n}(Z^{(k)} - \mathbb{E}[Z^{(k)}]) & \geq \alpha,\\
  \frac{1}{n} \sum_{k = 1}^{n}(Z^{(k)} - \mathbb{E}[Z^{(k)}]) & \geq \alpha(1 + \sigma^2).
\end{split}
\end{equation}

Since $X_i$ for all $i$ is approximately bounded, $Z^{(k)}$ is also approximately bounded. Let $Z^{(k)} \in [a_M, b_M]$ for all $k$ where $-\infty < a_M \leq b_M < +\infty$, then the probability of a crossover event is as follows using the Hoeffding's inequality (\ref{eq:7}):
\begin{equation}\label{eq:34}
  Pr\left(\frac{1}{n} \sum_{k = 1}^{n}(Z^{(k)} - \mathbb{E}[Z^{(k)}]) \geq \alpha(1 + \sigma^2)\right) \leq e^{-\frac{2n\left[\alpha(1 + \sigma^2)\right]^2}{(b_M - a_M)^2}},
\end{equation}
where $\sigma^2$ is the variance of Gaussian noise random variable $N$, and $\alpha = \frac{1}{1 + \sigma^2}(\rho_{ij} - \rho_{rs} + t)$.
\end{proof}

\section{Proof of Lemma \ref{lemma3}}\label{append3}
\begin{proof}
We know that the mutual information relation $I_e > I_{e'}$ holds for the edges $e$ and $e'$, and the crossover event will happen if Chow-Liu algorithm estimates the mutual information in reverse order, e.g., $\hat{I}_e \leq \hat{I}_{e'}$.

The correlation coefficients of approximately bounded Gaussian random variables $\varrho_{ij}$ and Gaussian random variables $\rho_{ij}$ have the following relation $\varrho_{ij} \approx -\rho_{ij}$ [Eq. 6 of \cite{Kugiumtzis}]. Therefore, we can use the correlation coefficient $\rho_{ij}$, and the correlation coefficient $\rho_{ij}$ is the sufficient statistics for mixed random variables $Y_i$ and $Y_j$, i.e., for Gaussian random variables and discrete random variables.

Let $\rho_e$ and $\rho_{e'}$ be the correlation coefficients for the edges $e$ and $e'$, respectively. Let $\hat{\rho}_e$ and $\hat{\rho}_{e'}$ be the estimators of the correlation coefficients of the edges. The crossover event is defined (using the positive correlation coefficient assumption) as:
\begin{equation}\label{eq:35}
  \hat{\rho}_{e'} - \hat{\rho}_e = \frac{1}{n} \sum_{k = 1}^{n}(Y_r^{(k)}Y_s^{(k)} - Y_i^{(k)}Y_j^{(k)}) \geq 0,
\end{equation}
where $Y_i = E_iX_i$ and $E_i$ is the random variable taking values in $\{1, ?\}$ representing an Erasure event.

Furthermore, the received symbol at the FC is represented by random variable $Y_i$ for sensor $i$. The Erasure event probability is defined as $\xi = \xi_i = Pr(Y_i = ? \not \in \mathcal{Y})$ where $\mathcal{Y}$ is the output alphabet, moreover, an Erasure is machine and time independent. \textbf{The Erased symbol $?$ is replaced with value $1$ for calculating sample correlation coefficients}.

Moreover, the approximately bounded assumption $|Y|\leq M\sigma$ where $M \geq 3$ is made, then using the same technique as for Part 2 in Lemma \ref{lemma2}, the probability of a crossover event is bounded using Hoeffding's inequality as follows:
\begin{equation}\label{eq:36}
  Pr\left(\frac{1}{n} \sum_{k = 1}^{n}(Z^{(k)} - \mathbb{E}[Z^{(k)}]) \geq \beta\right) \leq e^{-\frac{2n\beta^2}{(b_M - a_M)^2}},
\end{equation}
where $\beta = \rho_{e} - \rho_{e'}$ (difference of the correlation coefficients), $Z^{(k)} = Y_r^{(k)}Y_s^{(k)} - Y_i^{(k)}Y_j^{(k)}$, and $Z^{(k)} \in [a_M, b_M]$.
\end{proof}

\section{Proof of Lemma \ref{lemma4}}\label{append4}
\begin{proof}
Let machines $i$, $j$, $r$, and $s$ send the quantized data, represented by random variables $U_i$, $U_j$, $U_r$, and $U_s$ for all the samples $(k = 1, ..., n)$, respectively. Let a pair of edges be $e = (i, j)$ and $e' = (r, s)$. Due to the existence of binary symmetric channels between these machines and the FC, the received symbols at the FC are represented by $\hat{U}_i = R_i U_i$, $\hat{U}_j = R_j U_j$,
$\hat{U}_r = R_r U_r$, and $\hat{U}_s = R_s U_s$ where $\{R_i\}$ are i.i.d. Bernoulli random variables taking values $\{-1, +1\}$ with probability $\epsilon$ and are independent with the data random variable $U_i$.

Let the random variable $T^{(k)}$ be defined as
\begin{equation}\label{eq:37}
  \begin{split}
  T^{(k)} & = \mathcal{I}(\hat{U}_r^{(k)}\hat{U}_s^{(k)} = 1) - \mathcal{I}(\hat{U}_i^{(k)}\hat{U}_j^{(k)} = 1),\\
  & = \mathcal{I}(R_r R_s U_r^{(k)}U_s^{(k)} = 1) - \mathcal{I}(R_i R_j U_i^{(k)}U_j^{(k)} = 1),
  \end{split}
\end{equation}
where $\mathcal{I}(\cdot)$ is the indicator function. In addition, $T$ is the generic random variable of $T^{(k)}$. The probability of a crossover event can be written as
\begin{equation}\label{eq:38}
  \begin{split}
  Pr(\hat{\theta}_e \leq \hat{\theta}_{e'}) & = Pr\left(\sum_{k = 1}^n T^{(k)} \geq 0 \right),\\
  & = Pr\left(e^{\lambda \sum_{k=1}^{n} T^{(k)}} \geq 1\right), \quad \lambda > 0\\
  & \stackrel{(a)}{\leq} \mathbb{E}\left[e^{\lambda \sum_{k=1}^{n} T^{(k)}}\right], \\
  & = \left(\mathbb{E}\left[e^{\lambda T}\right]\right )^n,\\
  & = \left(p_0 + p_1 e^{\lambda} + p_2 e^{-\lambda}\right)^n,
  \end{split}
\end{equation}
where the inequality $(a)$  is by Markov's inequality, and the random variable $T^{(k)}$ can have values $\{0, 1, -1\}$ with probabilities $\{p_0, p_1, p_2\}$ defined in formulas (\ref{eq:20}-\ref{eq:22}). Chernoff bound is obtained by minimizing the last expression in formula (\ref{eq:38}) for $\lambda > 0$ as follows:
\begin{equation}\label{eq:39}
  Pr(\hat{\theta}_e \leq \hat{\theta}_{e'}) \leq (p_0 + 2\sqrt{p_1 p_2})^n = e^{n D},
\end{equation}
where $D = ln(p_0 + 2 \sqrt{p_1p_2}) \leq 0$.
\end{proof}

\section{Proof of Theorem \ref{theorem3}}\label{append5}
\begin{proof}
The formal proof of Algorithmic Bound (Algorithm \ref{algorithm2}) is given using the induction method and the impact of dominant errors (e.g., Definition \ref{def2}).

\noindent The upper bounds on the subtrees in array $A_1$ are in array $U_1$ (lines 5 in Algorithm \ref{algorithm2}). Similarly the upper bounds on the neighborhoods in array $A_2$ are in array $U_2$ (lines 10).

\noindent Take two subtrees $T_1, T_2 \in A_1$ and their corresponding upper bounds $u_1, u_2 \in U_1$. Suppose node $i \in T_1$ and node $j \in T_2$ (their neighborhoods $N(i), N(j) \in A_2$) are the potential nodes for connecting the potential edge $e_{ij}$, then using Definition \ref{def2}, we have computed the possible errors for this edge which are $v_i, v_j \in U_2$ (lines 10). Hence the total upper bound will be the sum $B = u_1 + u_2 + v_i + v_j$ (line 15).

\noindent In similar way, let $\hat{T}$ be the connected tree of $T_1, T_2$ with upper bound $B$. Take another subtree $T_3 \in A_1$ with upper bound $u_3 \in U_1$.
Let node $f \in \hat{T}$ and node $g \in T_3$ be the potential nodes for creating the potenital edge, and we also have the possible errors of this edge which are $v_f, v_g \in U_2$. Therefore, the upper bound will be $B + u_3 + v_f + v_g$.

\noindent Finally, continuation of the above argument shows that the upper bound in line 17 in Algorithm \ref{algorithm2} is actually the valid bound on the incorrect tree structure recovery probability.
\end{proof}

\ifCLASSOPTIONcaptionsoff
  \newpage
\fi


\begin{thebibliography}{1}


\bibitem{YWang}
Y.~Wang, X.~Li, and R.~Ruiz, ``Weighted general group lasso for gene selection in cancer classification,'' \textit{IEEE Trans. Cybernetics}, vol. 49, no. 8, pp. 2860-2873, Aug. 2019.

\bibitem{Murphy}
K.~P. Murphy, \textit{Machine learning: a probabilistic perspective}, MIT press, 2012.

\bibitem{Chow}
C.~Chow and C.~Liu, ``Approximating discrete probability distributions with dependence trees,'' \textit{IEEE Trans. Inf. Theory}, vol. 14, no. 3, pp. 462-467, May 1968.

\bibitem{Bresler}
G.~Bresler and M.~Karzand, ``Learning a tree structured Ising model in order to make predictions,'' \textit{The Annals of Statistics}, vol. 48, no. 2, pp. 713–737, Aug. 2020.

\bibitem{Drton}
M.~Drton and M.~H. Maathuis, ``Structure learning in graphical modeling,'' \textit{Annu. Rev. Stat. Appl.}, vol. 4, pp. 365-393, Mar. 2017.

\bibitem{SHuang}
S.~Huang \textit{et al.}, ``Learning brain connectivity of Alzheimer's disease by sparse inverse covariance estimation,'' \textit{NeuroImage}, vol. 50, no. 3, pp. 935-949, Apr. 2010.

\bibitem{RXiang}
R.~Xiang, J.~Neville, and M.~Rogati, ``Modeling relationship strength in online social networks,'' in \textit{Proc. of the 19th Int. Conf. World Wide Web}, Raleigh North Carolina, USA, Apr. 26-30, 2010, pp. 981-990.

\bibitem{Tan}
V.~Y. F. Tan, A.~Anandkumar, L.~Tong, and A.~S. Willsky, ``A large deviation analysis of the maximum likelihood learning of Markov tree structures,'' \textit{IEEE Trans. Inf. Theory}, vol. 57, no. 3, pp. 1714-1735, Mar. 2011.

\bibitem{F.Tan}
V.~Y. F. Tan, A.~Anandkumar, and A.~S. Willsky, ``Learning Gaussian tree models: analysis of error exponents and extremal structures,'' \textit{IEEE Trans. Signal Process.}, vol. 58, no. 5, pp. 2701-2714, May 2010.

\bibitem{Tavass}
M.~Tavassolipour, S.~A. Motahari, and M.~T. M Shalmani, ``Learning of tree structured Gaussian graphical models on distributed data under communication constraints,'' \textit{IEEE Trans. Signal Process.}, vol. 67, no. 1, pp. 17-28, Jan. 2019.

\bibitem{A.Tavass}
M.~Tavassolipour, A.~Karamzade, R.~Mirzaeifard, S.~A. Motahari, and M.~T. M Shalmani, ``Structure learning of sparse GGMs over multiple access networks,'' \textit{IEEE Trans. Communications}, vol. 68, no. 2, pp. 987-997, Feb. 2020.

\bibitem{Kang}
Z.~Kang, H.~Pan, S.~C. H. Hoi, and Z.~Xu, ``Robust graph learning from noisy data,'' \textit{IEEE Trans. Cybernetics}, vol. 50, no. 5, pp. 1833-1843, May 2020.

\bibitem{Katiyar}
A.~Katiyar, J.~Hoffmann, and C.~Caramanis, `` Robust estimation of tree structured Gaussian graphical models,'' in \textit{Proc. of the ICML 2019}, Long Beach, California, USA, June, 2019, pp. 3292–3300.

\bibitem{Katiyar2}
A.~Katiyar, V.~Shah, and C.~Caramanis, `` Robust estimation of tree structured Ising models,'' arXiv: 2006.05601v1 [stat.ML], June, 2020.

\bibitem{Jang}
H.~Jang, H.~S. Song, and Y.~Yi, ``Learning data dependency with communication cost,'' in \textit{Proc. of the Eighteenth ACM Int. Symposium on Mobile Ad Hoc Networking and Computing}, Los Angeles, CA, USA, June 26-29, 2018, pp. 171-180.

\bibitem{Wiesel}
A.~Wiesel and A.~O. Hero, ``Distributed covariance estimation in Gaussian graphical models,'' \textit{IEEE Trans. Signal Process.}, vol. 60, no. 1, pp. 211-220, Jan. 2012.

\bibitem{Meng}
Z.~Meng, D.~Wei, A.~Wiesel, and A.~O. Hero, ``Marginal likelihoods for distributed parameter estimation of Gaussian graphical models,'' \textit{IEEE Trans. Signal Process.}, vol. 62, no. 20, pp. 5425-5438, Oct. 2014.

\bibitem{Nikolakakis}
K.~E. Nikolakakis, D.~S. Kalogerias, and A.~D. Sarwate, ``Learning tree structures from noisy data,'' in \textit{Proc. of the 22nd Int. Conference on Artificial Intelligence and Statistics (AISTATS)}, Naha, Okinawa, Japan, vol. PMLR 89, Apr. 16-18, 2019, pp. 1771-1782.

\bibitem{Nikolakakis1}
K.~E. Nikolakakis, D.~S. Kalogerias, and A.~D. Sarwate, ``Predictive learning on hidden tree-structured Ising models,'' \textit{Jour. of Machine Learning Research}, vol. 22, no. 59, pp. 1-82, Feb. 2021.

\bibitem{Nikolakakis2}
K.~E. Nikolakakis, D.~S. Kalogerias, and A.~D. Sarwate, ``Optimal rates for learning hidden tree structures,'' arXiv:1909.09596v4 [stat.ML] Mar. 2021.

\bibitem{Tandon2}
A.~Tandon, A.~J. Y. Han, and V.~Y. F. Tan, ``SGA: a robust algorithm for partial recovery of tree-structured graphical models with noisy samples,'' in \textit{Proc. of the 38th Int. Conference on Machine Learning}, vol. PMLR 139, July 18-24, 2021, pp. 10107-10117.

\bibitem{Yu}
C.~Yu, H.~Tang, C.~Renggli, S.~Kassing, A.~Singla, D.~Alistarh, C.~Zhang, and J.~Liu, ``Distributed learning over unreliable networks,'' in \textit{Proc. of the 36th Int. Conference on Machine Learning}, Long Beach, California, USA, vol. PMLR 97, June 9-15, 2019, pp. 7202-7212.

\bibitem{Tandon}
A.~Tandon, V.~Y. F. Tan, and S.~Zhu, ``Exact asymptotics for learning tree-structured graphical models with side information: noiseless and noisy Samples,'' \textit{IEEE Journal on selected areas Info. Theory}, vol. 1, no. 3, pp. 760-776, Nov. 2020.

\bibitem{Taskar}
B.~Taskar, V.~Chatalbashev, and D.~Koller, ``Learning associative Markov networks,'' in \textit{Proc. of the 21st Int. Conference on Machine Learning}, Banff, Alberta, Canada, 2004, pp. 1-10.

\bibitem{Krumsiek}
J.~Krumsiek, K.~Suhre, T.~Illig, J.~Adamski, and F.~J Theis, ``Gaussian graphical modeling reconstructs pathway reactions from high-throughput metabolomics data,'' \textit{BMC Systems Biology}, vol. 5, no. 21, pp. 1-16, Jan. 2011.

\bibitem{Bishop}
C.~M. Bishop, \textit{Pattern recognition and machine learning},  Springer-Verlag New York, 2006.

\bibitem{Kruskal}
J.~B. Kruskal, ``On the shortest spanning subtree of a graph and the traveling salesman problem,'' \textit{Proc. of the American Mathematical Society}, vol. 7, no. 1, pp. 48-50, Feb. 1956.

\bibitem{Prim}
R.~C. Prim, ``Shortest connection networks and some generalizations,'' \textit{The Bell System Technical Journal}, vol. 36, no. 6, pp. 1389-1401, Nov. 1957.

\bibitem{Kugiumtzis}
D.~Kugiumtzis and E.~B. Senta, ``Normal correlation coefficient of non-normal variables using piece-wise linear approximation,'' \textit{Comput Stat}, vol. 25, no. 4, pp. 645-662, Apr. 2010.

\bibitem{Ware}
R. Ware and F. Lad, ``Approximating the distribution for sums of product of normal variables,'' Research-Report, \textit{Department of Mathematics and Statistics, University of Canterbury, New Zealand}, 2003.

\bibitem{Macias}
A.~S. Macıas, ``An approach to distribution of the product of two normal variables,'' \textit{Discussiones Mathematicae Probability and Statistics}, vol. 32, pp. 87-99, 2012.

\bibitem{Lee}
K.~Lee, M.~Lam, R.~Pedarsani, D.~Papailiopoulos, and K.~Ramchandran, ``Speeding up distributed machine learning using codes,'' \textit{IEEE Trans. Inf. Theory}, vol. 64, no. 3, pp. 1514-1529, Aug. 2017.

\bibitem{Buyukates}
B.~Buyukates and S.~Ulukus, ``Timely distributed computation with stragglers,'' \textit{IEEE Trans. Communications}, vol. 68, no. 9, pp. 5273-5282, Sep. 2020.

\bibitem{Gamal}
M.~EL Gamal and L.~Lai, ``On rate requirements for achieving the centralized performance in distributed estimation,'' \textit{IEEE Trans. Signal Process.}, vol. 65, no. 8, pp. 2020-2032, Apr. 2017.


\bibitem{Xiao}
J.~J. Xiao, S.~Cui, Z.~ Q. Luo, and A.~J. Goldsmith, ``Linear coherent decentralized estimation,'' \textit{IEEE Trans. Signal Process.}, vol. 56, no. 2, pp. 757-770, Feb. 2008.

\end{thebibliography}
\end{document}